\documentclass[10pt,twocolumn,letterpaper]{article}

\usepackage[accsupp]{axessibility}
\usepackage{iccv}

\definecolor{iccvblue}{rgb}{0.21,0.49,0.74}
\usepackage[pagebackref,breaklinks,colorlinks,allcolors=iccvblue]{hyperref}

\usepackage{algorithm}
\usepackage{multirow}
\usepackage{algpseudocode}
\usepackage{xcolor}
\usepackage{fontawesome5}
\usepackage{soul}
\def\paperID{5364} %
\def\confName{ICCV}
\def\confYear{2025}

\title{Trokens: Semantic-Aware Relational Trajectory Tokens 
\\ for Few-Shot Action Recognition}

\author{
Pulkit Kumar$^{1}$\thanks{Equal Contribution} \quad Shuaiyi Huang$^{1*}$ \\ \quad Matthew Walmer$^1$ \quad Sai Saketh Rambhatla$^{1,2}$ \quad Abhinav Shrivastava$^1$\\
\normalsize
$^1$University of Maryland, College Park \qquad $^2$GenAI, Meta\\
{\tt \footnotesize \{pulkit,huangshy,mwalmer,abhinav\}@cs.umd.edu} {\tt \footnotesize \ rssaketh@meta.com} 
}

 \xspaceaddexceptions{]\}}
\def\ours{Trokens\xspace}

\def\cls{\texttt{CLS}\xspace}
\usepackage{bbding}

\newcommand{\tick}{\Checkmark}
\newcommand{\cross}{\XSolidBrush}

\begin{document}
\maketitle

\begin{abstract}

Video understanding requires effective modeling of both motion and appearance information, particularly for few-shot action recognition. While recent advances in point tracking have been shown to improve few-shot action recognition, two fundamental challenges persist: selecting informative points to track and effectively modeling their motion patterns. We present \ours, a novel approach that transforms trajectory points into semantic-aware relational tokens for action recognition. First, we introduce a semantic-aware sampling strategy to adaptively distribute tracking points based on object scale and semantic relevance. Second, we develop a motion modeling framework that captures both intra-trajectory dynamics through the Histogram of Oriented Displacements (HoD) and inter-trajectory relationships to model complex action patterns. Our approach effectively combines these trajectory tokens with semantic features to enhance appearance features with motion information, achieving state-of-the-art performance across six diverse few-shot action recognition benchmarks: Something-Something-V2 (both full and small splits), Kinetics, UCF101, HMDB51, and FineGym. Our project page is available \href{https://trokens-iccv25.github.io}{here}.

\end{abstract}

\section{Introduction}
\begin{figure}[t!]
    \centering
    \includegraphics[width=\columnwidth]{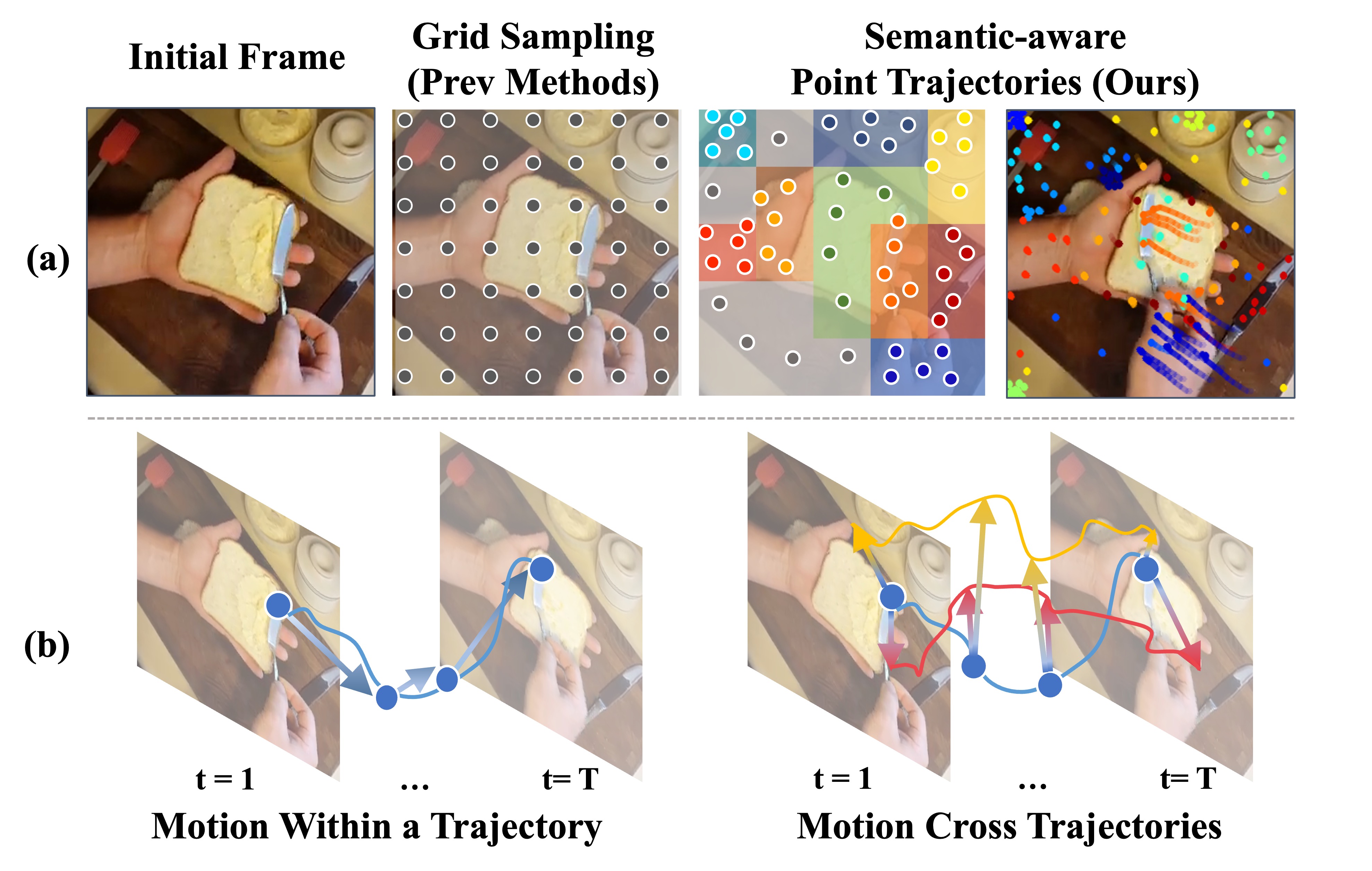}
    \caption{\textbf{Our motivation.} (a) Our semantic-aware points adapt better to object scale and semantic relevance while existing methods with grid sampling miss small objects with important motion (e.g., knife). (b) We explicitly model relational motions within a trajectory and across trajectories.}
    \label{fig:motivation}
    \vspace{-3mm}
\end{figure}

At the core of video understanding lies the fundamental synergy between motion and appearance cues. Motion patterns reveal the dynamic flow of actions through time. At the same time, appearance information captures the rich context, the interplay of objects, environments, and their relationships within each frame. In action recognition tasks, particularly in few-shot settings, explicitly modeling this complementary relationship becomes crucial, as both aspects provide distinct yet essential signals.

While appearance understanding in models has made great strides, the challenge of capturing crucial motion patterns remains complex. Traditional optical flow techniques \cite{laptev2008learning, pervs2010histograms, simonyan2014two} have been a primary approach, but they are fundamentally limited to analyzing adjacent frames and deteriorate under occlusions, resulting in incomplete motion representations. In parallel to optical flow methods, trajectory-based approaches emerged as an alternative paradigm. Early trajectory works \cite{wang2015action, wang2013dense, wang2013action} made progress by capturing longer-term patterns, and recent work \cite{tats} has further advanced this direction through point tracking. Unlike optical flow, point tracking~\cite{cotracker,locotrack,alltracker,tapnext} explicitly maintains temporal correspondence across long sequences and handles occlusions naturally, making it particularly effective for capturing complex motion patterns in real-world actions. In this work, we aim to advance few-shot action recognition by building upon these advantages of point tracking approaches.

To advance point tracking for action recognition, we must address two fundamental challenges: (1) sampling informative query points to track through time, and (2) effectively modeling the complex motion patterns captured by these trajectories. Dense point sampling provides comprehensive coverage but is computationally expensive, while sparse sampling risks missing crucial motion information, especially from smaller objects essential for action understanding. Beyond these tracking challenges, current transformer-based approaches~\cite{arnab2021vivit,timesformer,patrick2021keeping} attempt to learn motion patterns implicitly for action recognition, but it remains unclear whether these models truly capture and leverage motion information or rely on other contextual cues.

To address these limitations, we identify two key questions: (1) how can we develop an effective sampling strategy that balances coverage and efficiency?~and (2) how can we explicitly model and utilize the motion patterns captured in point trajectories? In this paper, we present a novel approach that leverages both semantic-aware sampling and explicit motion modeling to improve point tracking-based video understanding.

For the first challenge of effective point sampling, we propose a semantic-aware sampling strategy that adapts to object scale and importance. This approach is particularly crucial for actions involving small but critical objects, as illustrated in Figure 1(a), where the knife spreading the butter could be easily missed by uniform sampling due to its small size. By leveraging semantic information extracted from DINO~\cite{dinov2} patch tokens, our method ensures comprehensive coverage of action-relevant objects regardless of scale. The sampling density, guided by semantic understanding, allocates denser points to smaller, action-critical objects (\textit{e.g.,} knife) and sparser sampling to larger regions (\textit{e.g.,} background desk). This approach ensures we capture the motion of all semantically meaningful objects while maintaining computational efficiency.

With our sampling strategy in place, we address the second challenge through our Relational Motion Modeling module, which processes point trajectories from modern trackers \cite{cotracker,locotrack,alltracker,tapnext} in two complementary ways. Revisiting the knife spreading butter example in Figure~\ref{fig:motivation}, our first component captures intra-trajectory dynamics using Histogram of Oriented Displacements (HoD)~\cite{hod} to model the knife's own movement patterns and directions. Our second component extracts inter-trajectory relationships by tracking how different objects (\textit{e.g.,} the knife and bread) interact with each other, revealing the distinctive motion patterns that define actions. This Relational Motion Modeling module effectively captures both individual object movements and their meaningful interactions (Figure 1(b)).

Overall, we introduce~\ours, a novel framework that leverages semantic-aware sampling and explicit motion modeling to effectively bridge motion and appearance information. Our contributions are as follows:
\begin{itemize}
\item We propose a novel semantic-aware point sampling approach leveraging semantic priors to adaptively distribute tracking points based on object scale and relevance.
\item We develop a novel Relational Motion Modeling module that explicitly captures both intra- and inter-trajectory dynamics to understand complex motion patterns.
\item We conduct comprehensive experiments with \ours on six few-shot action recognition benchmarks including Something-Something (Full \& Small)~\cite{ssv2}, Kinetics~\cite{kinetics}, UCF101~\cite{ucf}, HMDB51~\cite{hmdb}, and FineGym~\cite{finegym}, and achieve state-of-the-art performance.
\end{itemize}

\section{Related Work}

\paragraph{Few-Shot Action Recognition.}
Human Action Recognition requires one to model the complex temporal dynamics of a scene while also filtering out the redundant information shared between frames \cite{kliper2011one, poppe2010survey, wang2021proposal, wang2021self}. While these challenges are typically addressed using large training datasets, in the setting of few-shot action recognition, methods must instead use well-constructed mechanisms to achieve effective performance with limited data. Many methods are based on metric-learning \cite{otam, trx, hyrsm, hyrsm++, hcl, bishay2019tarn, zhang2021learning, zhang2020few, wang2021semantic, ni2022multimodal, fu2020depth, tran2015learning}, introducing various mechanisms to determine if two videos are similar or different, thus enabling action classification. Meanwhile, other methods focus on improving feature representations for spatio-temporal modeling \cite{strm, sloshnet, hyrsm, mtfan, zhu2018compound, zhu2020label, gghm, molo, tats}.
Some recent works also leverage multi-modal language-image pretraining \cite{radford2021learning} and/or additional text data at training or inference time to further enhance performance \cite{wang2024clip, wu2024efficient, cao2024exploring, deng2024text, zhu2017structured, wang2024sfmm, li2024frame, guo2024multi, tang2024semantic, cao2024task}. While these works show strong results, they represent a bifurcation of the field into two domains: multi-modal and vision-only few-shot action recognition. Our method, \ours, is a vision-only method, and we focus on primarily comparing with like baselines.  A recent work~\cite{manta} proposes a state space based architecture for long sequence few shot action recognition. While promising, improving the core architecture is orthogonal to our contributions.

\vspace{-3mm}
\paragraph{Point Tracking for Feature Learning.}
Point tracking has a long history in computer vision research, and now recent advances in the field have enabled the efficient generation of dense and high-quality point tracks \cite{doersch2022tap, harley2022particle, doersch2023tapir, zheng2023pointodyssey, huang2019dynamic, huang2020confidence, he2023towards, huang2022learning, huang2024uvis, huang2024point, moing2023dense, cotracker, wang2023tracking, dino_tracker}. 
These tracks lend themselves well to a fundamental element of video learning: the disentanglement of motion and appearance information. Several prior works have successfully applied point tracks to guide the extraction of deep and classical features \cite{wang2015action, huang2024ardup, zheng2024tracevla, wu2024autohallusion, huang2025trend, wang2013dense, wang2013action}. 
The recent work TATs \cite{tats} further demonstrates the power of point tracking for transformer token pooling in few-shot action recognition.
However, TATs falls short on our two key challenges: its uniform grid-based sampling fails to adapt to object scales, and it treats trajectories merely as feature anchors, neglecting the rich motion patterns they contain. These limitations motivate our \ours approach that addresses both sampling and motion modeling challenges.

\paragraph{Motion Features.}

In recent years, many architectures and approaches have been proposed to learn joint or disentangled appearance and motion features from video \cite{wang2018appearance, zhao2018recognize, kwon2020motionsqueeze, wang2020video, zhang2023extracting}.
However, such approaches are reliant on training data and struggle in low-data few-shot regimes.
Meanwhile, other methods have been proposed to model motion features directly from optical flow \cite{laptev2008learning, pervs2010histograms}, or point trajectories \cite{wang2013dense, abdul2015human, wang2016action}.
In this work, we aim to improve few-shot action recognition performance by efficiently leveraging motion features derived from our trajectories.
We draw inspiration from classical vision methods like Histogram of Oriented Gradients (HoG) \cite{dalal2005histograms}, and Histogram of Oriented Displacements (HoD) \cite{hod}. Specifically, we present a new implementation of HoD tailored toward general object intra-track motion features, which we describe in Section \ref{subsec:method_motion}. While the original HoD is focused on only human skeleton keypoints, our version is designed for general object motion characterization. Additionally, we preserve temporal order through per-timestep computation rather than whole-trajectory pyramidal aggregation like \cite{hod}.

\begin{figure*}[t!]
    \centering
    \includegraphics[trim={6.0cm 0 0 0}, width=\textwidth]{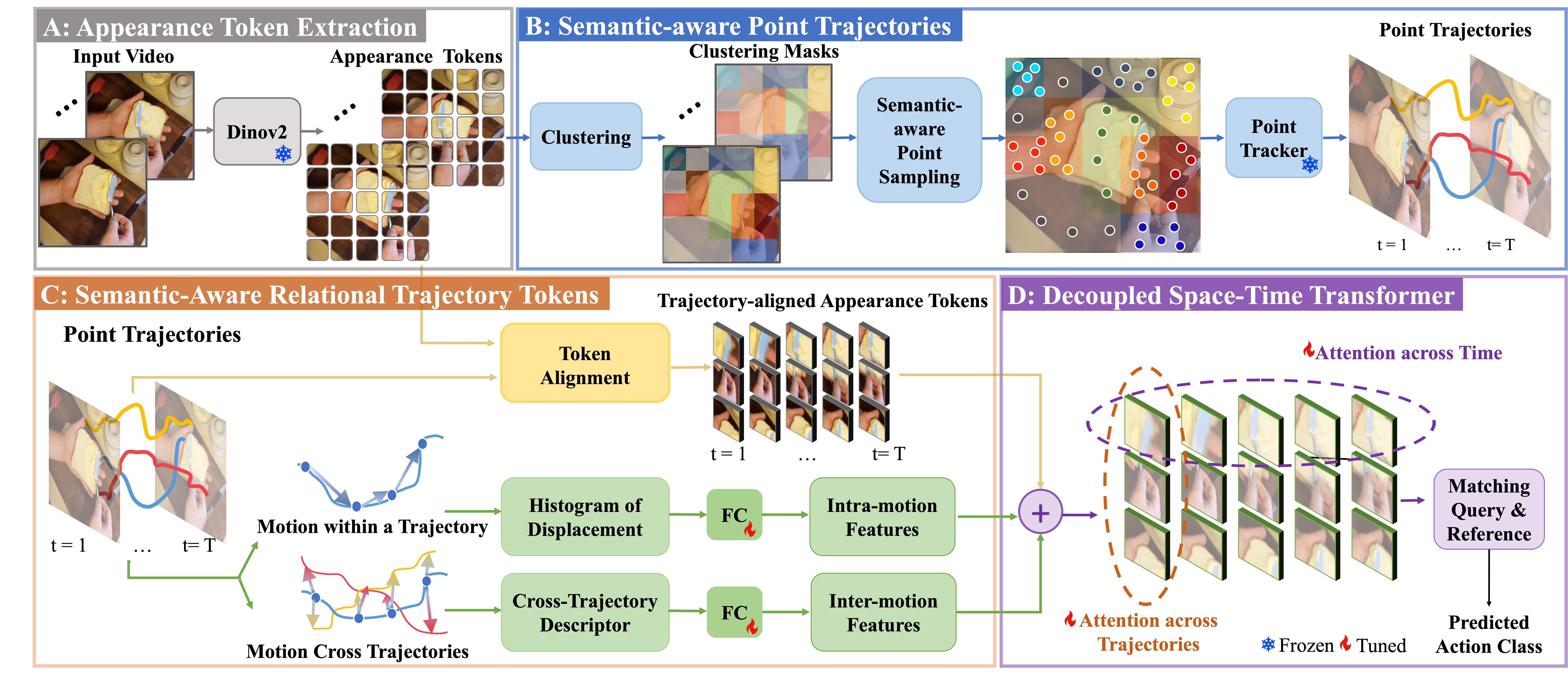}
    \caption{\textbf{Method Overview.} \textcolor{gray}{(A)} Given an input video, we extract appearance tokens using DINOv2. \textcolor{blue}{(B)} We then cluster these tokens and sample semantic-aware points in the initial frame, which are tracked using Co-tracker~\cite{cotracker} to obtain point trajectories. \textcolor{orange}{(C)} We compute intra- and inter-motion features, reorder appearance tokens via token alignment~\cite{tats}, and fuse them with motion features via element-wise addition to form semantic-aware relational trajectory tokens. \textcolor{DarkOrchid}{(D)} Finally, we input these tokens into a Decoupled Space-Time Transformer for few-shot action classification.}
    \label{fig:yourlabel}
    \vspace{-3mm}
\end{figure*}

\section{Few-shot Action Recognition Setup}

Few-shot action recognition is the problem of recognizing novel action classes with few labeled instances per class. Unlike fully supervised learning, training and test classes are mutually exclusive in the few-shot setting. Formally, given a training set $D_{\text{train}} = \{(v_i, y_i) \mid y_i \in C_{\text{train}}\}$ and test set $D_{\text{test}} = \{(v_i, y_i) \mid y_i \in C_{\text{test}}\}$,  $C_{\mathrm{train}} \cap C_{\mathrm{test}} = \phi$. Training is done with episode-based meta-learning where each episode has a support set $S$ with $N$ classes and $K$ examples per class (termed $N$-way $K$-shot, e.g., 5-way 1-shot), and a query set $Q$ with samples to classify into these $N$ classes. 

\section{Our Approach}

\subsection{Overview}

Our work aims to leverage semantic and motion priors for few-shot video action classification. We propose the following key components: (1) Semantic-aware point trajectories sampling, where we leverage DINO features to sample semantically meaningful points for motion tracking (Sec~\ref{subsec:semantic}); and (2) a Relational Motion Modeling module that explicitly models motion changes within individual trajectories and across trajectories to capture detailed movement patterns (Sec~\ref{subsec:method_motion}). 
The appearance features and the proposed motion features are sampled using the semantic-aware point trajectories and subsequently fused to form the trajectory-aligned tokens following~\cite{tats}.
 Next, we adopt a Decoupled Space-Time Transformer~\cite{tats}, to process the trajectory-aligned tokens. 
All components of our method are trained in an end-to-end fashion using a standard few-shot loss~\cite{molo}.

\subsection{Semantic-aware Point Trajectories}
\label{subsec:semantic}

Dense point tracking~\cite{cotracker,cotracker3} offers promising video understanding capabilities, but its effectiveness depends on the initial selection of tracking points. The standard practice for point-based few-shot action classification is to sample points in a uniform grid~\cite{tats}. Employing uniform grid sampling, despite its simplicity, often under-samples small objects crucial for understanding actions while capturing redundant information from background regions.

To address this limitation, we leverage a semantic prior to guide point selection. DINO's self-supervised learning framework produces patch tokens with rich semantic information, where tokens from the same object naturally cluster in feature space~\cite{hamilton2022unsupervised, LOST,Rambhatla2023MOSTMO, Wang2023CutAL, Wang2023VideoCutLERSS}. Leveraging this property, we construct semantic-aware clustering masks over patch tokens and sample points accordingly. Our strategy enables semantic-aware point sampling that adapts to object scale and semantic relevance.

Formally, we extract DINO appearance features $\mathcal{F}^{\text{RGB}}\in\mathcal{R}^{H\times W\times T \times C}$ from the input video, where $T$ is the number of frames and $H,W$ are spatial dimensions. We cluster the appearance features~\cite{bolya2022token} into $L$ groups which are subsequently used to sample semantic-aware points.
For $M$ trajectories, we sample $q = \frac{M}{L}$  points per cluster, from the first frame where a new semantic cluster appears, as the semantic-aware points. We denote these points for all clusters as $P_s = \{(x_s^i,y_s^i)\}_{i=1}^{M}$, where ($x_i, y_i$) is the spatial coordinate of a point. These points serve as initialization for trajectory extraction. 

To track the sampled points, we utilize pretrained dense point tracking model Co-tracker~\cite{cotracker} (denoted as $\mathcal{T}$). The extracted point trajectories, known as semantic-aware point trajectories, are given by $\mathcal{P} = \mathcal{T}(P_s) = \{ \mathcal{P}^{m} \}_{m=1}^{M}$. Each trajectory \( \mathcal{P}^{m} = \left[(x_{t}^{m}, y_{t}^{m})\right]_{t=1}^{T}\in \mathcal{R}^{T\times 2} \), captures the motion of a point over T frames.  As shown in Fig.~\ref{fig:motivation}, our approach provides better coverage of small but significant objects while reducing redundant background trajectories compared to uniform sampling.

\subsection{Relational Motion Modeling}
\label{subsec:method_motion}

Given semantic-aware point trajectories, a natural question is how to best capture their rich motion dynamics. To this end, we propose to explicitly model motion in two key aspects: dynamics within individual trajectories (intra-motion) and relationships across different trajectories (inter-motion). Such fine-grained representations capture both local motion patterns and cross trajectory interactions, providing discriminative features crucial for action recognition.

\paragraph{Intra-motion Module.}
Inspired by the Histogram of Oriented Gradients (HoG)~\cite{dalal2005histograms}, we revisit \textit{Histogram of Displacement (HoD)}~\cite{gowayyed2013histogram} to encode both magnitude and orientation changes over time on top of point trajectories. 
Given a trajectory $\mathcal{P}^m$, we compute the displacement at time $t$ as \( \Delta x_t = (x_t - x_{t-\delta}) \) and \( \Delta y_t = (y_t - y_{t-\delta}) \), where \( \delta \) is a hyperparameter controlling the temporal interval (we omit $m$ for simplicity). The displacement magnitude is \( \Delta{d}_t = \sqrt{\Delta x_t^2 + \Delta y_t^2} \), and the direction is \( \theta_t = \arctan2(\Delta y_t, \Delta x_t) \), with zero padding for $t < \delta$.
For each timestep, we bin the orientation $\theta_t$ into a histogram with $B$ bins spanning 360 degrees (\textit{e.g.,} $B=32$). Each displacement \( \Delta{d}_t \) contributes to the two nearest orientation bins proportionally, weighted by its magnitude. This produces a histogram of displacement descriptor for each trajectory:
\[
\mathbf{H}_{\text{HoD}} = f_\text{HoD}(\mathcal{P}^m) \in \mathbb{R}^{T \times B}
\]

\noindent
We then apply a fully connected layer to project this descriptor into a $C$-dimensional space for all $M$ trajectories to obtain our intra-motion features $\mathcal{F}_{\text{intra}}^{\text{motion}}$ as below:
\[
\mathcal{F}_{\text{intra}}^{\text{motion}} = {\text{FC}}(f_\text{HoD}(\mathcal{P})) \in \mathbb{R}^{M \times T \times C}.
\]

\noindent
Our approach differs from prior work~\cite{gowayyed2013histogram} in encoding HoD in several aspects. First, we preserve temporal order through per-timestep computation rather than whole-trajectory aggregation. Second, we enhance expressiveness via learnable projections. Moreover, our formulation generalize beyond human keypoints to arbitrary trajectories for broader applicability to motion analysis tasks.

\paragraph{Inter-motion Module.}
While intra-motion features capture dynamics within individual trajectories, complex actions involve coordinated movements (e.g., relative hand positions distinguish ``opening a door'' from ``chopping vegetables''). We complement our intra-motion features with inter-motion modeling that captures evolving spatial relationships among trajectories.
Specifically, we compute pairwise relative displacements between trajectories.
For each trajectory \( \mathcal{P}^{m} = \left[(x_{t}^{m}, y_{t}^{m})\right]_{t=1}^{T} \) at time $t$, we define its cross-trajectory descriptor as:
\[
\mathbf{d}_{t}^{m} = \left[ (x_{t}^{m} - x_{t}^{m'}, y_{t}^{m} - y_{t}^{m'}) \right]_{m'=1}^{M} \in \mathbb{R}^{2M}
\]
This captures relative positions between trajectory $m$ and all other trajectories in a fixed order. The complete cross-trajectory descriptor $\mathbf{d} \in \mathbb{R}^{M \times T \times 2M}$ represents spatial relationships across all trajectories and timesteps.
Finally, we obtain inter-motion features by projecting the cross-trajectory descriptors to the feature space:
\[
\mathcal{F}_{\text{inter}}^{\text{motion}} = \text{FC}(\mathbf{d}) \in \mathbb{R}^{M \times T \times C},
\]
\noindent
It is worth noting that while transformers could potentially learn motion patterns through self-attention, their position embeddings primarily encode static locations without directly capturing temporal displacements or cross-trajectory relationships. Our explicit modeling of motion dynamics provides prior knowledge that helps the model focus on discriminative motion features rather than relying on self-attention to implicitly discover these patterns.

\subsection{Motion-aware Space-Time Transformer}

\label{subsec:tnf}
Given point trajectories $\mathcal{P}$, intra-, inter-motion features $\mathcal{F}_\text{intra}^{\text{motion}}$, $\mathcal{F}_\text{inter}^{\text{motion}}$, and appearance tokens from DINO $\mathcal{F}^{\text{RGB}}$, we utilize a transformer for spatiotemporal modeling of both motion and appearance. 
We first construct trajectory-aligned appearance tokens from point trajectories and appearance features following~\cite{tats}. Given video appearance tokens 
$\mathcal{F}^\text{RGB} \in \mathbb{R}^{H \times W \times T \times C}$
and point trajectories 
$\mathcal{P} \in \mathbb{R}^{M \times T \times 2}$, 
we extract trajectory-aligned appearance tokens:
\begin{equation}
\mathcal{F}_{\text{traj}}^{\text{RGB}} = \operatorname{Align}(\mathcal{F}^\text{RGB}, \mathcal{P}) \in \mathbb{R}^{M \times T \times C},
\end{equation}
where $\operatorname{Align}(\cdot)$ samples appearance features given point coordinates. This reordering aligns visual information with motion paths, helping self-attention learn motion explicitly.
We then fuse our intra-inter motion features and trajectory-aligned appearance tokens via element-wise addition, enabling the transformer to  capture both intra- and inter-trajectory dependencies:
\begin{equation}
\mathcal{F}^{\text{fuse}} = \mathcal{F}_{\text{traj}}^{\text{RGB}} + \mathcal{F}_{\text{intra}}^{\text{motion}} + \mathcal{F}_{\text{inter}}^{\text{motion}}.
\end{equation}
Given the semantic-aware relational trajectory tokens $\mathcal{F}^{\text{fuse}}$, we employ a decoupled attention strategy that processes the temporal and spatial dimensions separately following~\cite{tats}. Self-attention is applied within each trajectory to model temporal dependencies, and across trajectories to capture spatial relationships in parallel, the results of which are then added together to form the final output embeddings $\mathcal{F}_{\text{final}}$.

\noindent
Finally, we conduct cross-attention between a learnable \cls token and the final output embeddings $\mathcal{F}_{\text{final}}\in \mathbb{R}^{M \times T \times C}$, and produced the output class token $\mathbf{c}_{\text{cls}} \in \mathcal{R}^{C}$.

\subsection{Few-shot Loss}
\label{subsec:loss}

We extract the output embeddings $\mathcal{F}_{\text{final}}$ and class token $\mathbf{c}_{\text{cls}}$ for all samples in the support and query sets. To obtain the classification output, we apply a fully connected layer on the output class token, mapping it to the number of classes $C_{\text{train}}$ in the training set with $p_{\text{cls}} = \operatorname{FC}(\mathbf{c}_{\text{cls}}) \in \mathbb{R}^{C_{\text{train}}}$.

Following prior works~\cite{molo,hyrsm,hyrsm++,tats}, our loss consists of two parts. One is a cross-entropy loss applied to the classification output on the query set  $p_{\text{cls}}^{Q}$, capturing global $C_\text{train}$ class information. The other is a contrastive loss~\cite{molo} applied to final embeddings between the query set $\mathcal{F}_{\text{final}}^{\text{Q}}$ and the reference set $\mathcal{F}_{\text{final}}^{\text{S}}$ for $N$-way few-shot classification to encourage feature discrimination:
\begin{equation}
\mathcal{L} = \mathcal{L}_{\text{CE}}(p_{\text{cls}}^{Q}, y) + \mathcal{L}_{\text{Contrastive}}(\mathcal{F}_{\text{final}}^{Q}, \mathcal{F}_{\text{final}}^{S}).
\end{equation} We refer readers to \cite{tats} and \cite{molo} for further details.

\section{Experiments}
\subsection{Datasets} 
\label{sec:datasets}

We evaluate our approach's effectiveness across multiple action recognition benchmarks using established few-shot  splits: Something-Something~\cite{ssv2}, Kinetics~\cite{kinetics}, UCF101~\cite{ucf}, HMDB51~\cite{hmdb}, and FineGym~\cite{finegym}. For Something-Something, we use two standard configurations~\cite{otam}: SSV2 Small (100 samples per class; 100 classes) and SSV2 Full (all classes). Our evaluations follow the split protocols from previous works: Kinetics splits from~\cite{zhu2018compound}, UCF101 and HMDB51 splits from~\cite{mtfan,zhang2020few}, and FineGym splits from~\cite{tats}. To ensure fair comparison, we maintain consistency with the evaluation protocols used in prior works~\cite{tats,gghm,molo,hyrsm,hyrsm++}.
\subsection{Implementation details} 
\label{sec:implementation details}

We follow prior work~\cite{molo,tats} for most architectural choices and training configurations. Using DINOv2-base~\cite{dinov2}, we get semantic clusters from which 256 semantic-aware points are sampled for tracking via CoTracker~\cite{cotracker,cotracker3}. The architecture employs a single transformer block and uses 32-bin Histogram of Directions (HoD) in the intra-motion module. During training, only the transformer and motion modules are optimized while other components remain frozen. Following standard protocols~\cite{gghm,molo,hyrsm,tats}, we evaluate using average few-shot accuracy across 10,000 episodes. 

\subsection{Quantitative Results}

\begin{table*}[t]
\addtolength{\tabcolsep}{2pt} %

\caption{Comparison of few-shot action accuracy (1-5 shots) on SSV2 Full and Kinetics datasets versus contemporary methods. Best results are bolded, second-best underlined, and "-" indicates unavailable data.}
\label{table:full_kinetics}
\resizebox{\textwidth}{!}{

\begin{tabular}{@{}llccccc|ccccc@{}}

\toprule

& & \multicolumn{5}{c}{\textbf{SSV2 Full}} & \multicolumn{5}{c}{\textbf{Kinetics}}  \\ \cmidrule(l){3-7} \cmidrule(l){8-12}

\textbf{Method} & \textbf{Reference} & \multicolumn{1}{l}{1-shot} & \multicolumn{1}{l}{2-shot} & \multicolumn{1}{l}{3-shot} & \multicolumn{1}{l}{4-shot} & \multicolumn{1}{l}{5-shot} & \multicolumn{1}{l}{1-shot} & \multicolumn{1}{l}{2-shot} & \multicolumn{1}{l}{3-shot} & \multicolumn{1}{l}{4-shot} & \multicolumn{1}{l}{5-shot} \\
\midrule
OTAM~\cite{otam} & CVPR'20 & 42.8 & 49.1 & 51.5 & 52.0 & 52.3 & 72.2 & 75.9 & 78.7 & 81.9 & 84.2 \\
TRX~\cite{trx} & CVPR'21 & 42.0 & 53.1 & 57.6 & 61.1 & 64.6 & 63.6 & 76.2 & 81.8 & 83.4 & 85.2 \\
STRM~\cite{strm} & CVPR'22 & 43.1 & 53.3 & 59.1 & 61.7 & 68.1 & 62.9 & 76.4 & 81.1 & 83.8 & 86.7 \\
MTFAN~\cite{mtfan} & CVPR'22 & 45.7 & - & - & - & 60.4 & 74.6 & - & - & - & 87.4 \\
HYRSM~\cite{hyrsm} & CVPR'22 & 54.3 & 62.2 & 65.1 & 67.9 & 69.0 & 73.7 & 80.0 & 83.5 & 84.6 & 86.1 \\
HCL~\cite{hcl} & ECCV'22 & 47.3 & 54.5 & 59.0 & 62.4 & 64.9 & 73.7 & 79.1 & 82.4 & 84.0 & 85.8 \\
Nguyen et al~\cite{nguyen2022inductive} & ECCV'22 & 43.8 & - & - & - & 61.1 & 74.3 & - & - & - & 87.4 \\
Huang et al~\cite{huang2022compound} & ECCV'22 & 49.3 & - & - & - & 66.7 & 73.3 & - & - & - & 86.4 \\
MoLo~\cite{molo} & CVPR'23 & 56.6 & 62.3 & 67.0 & 68.5 & 70.6 & 74.0 & 80.4 & 83.7 & 84.7 & 85.6 \\
SloshNet~\cite{sloshnet} & AAA1'23 & 46.5 & - & - & - & 68.3 & 70.4 & - & - & - & 87.0 \\
GgHM~\cite{gghm} & ICCV'23 & 54.5 & - & - & - & 69.2 & 74.9 & - & - & - & 87.4 \\
RFPL~\cite{rfpl} & ICCV'23 & 47.0 & 54.6 & 58.3 & 60.3 & 61.0 & 74.6 & 80.0 & 82.1 & 84.1 & 86.8 \\
CCLN~\cite{ccln} & PAMI'24 & 46.0 & - & - & - & 61.3 & 75.8 & 82.1 & 85.0 & 86.1 & 87.5 \\
HYRSM++~\cite{hyrsm++} & PR'24 & 55.0 & 63.5 & 66.0 & 68.8 & 69.8 & 74.0 & 80.8 & 83.9 & 85.3 & 86.4 \\
TATS~\cite{tats} & ECCV'24 & \underline{57.7} & \underline{67.1} & \underline{70.0} & \underline{70.6} & \underline{74.6} & 81.9 & \underline{86.5} & \underline{89.9} & \underline{90.6} & 91.1 \\
TEAM~\cite{Lee_2025_CVPR} & CVPR'25 & - & - & - & - & - & \textbf{83.3} & - & - & - & \textbf{92.9} \\
\midrule
\textbf{\ours} & - & \textbf{61.5} & \textbf{69.9} & \textbf{73.8} &\textbf{75.9} & \textbf{76.7} & \underline{82.9} & \textbf{87.7} & \textbf{89.9 }& \textbf{90.8} & \underline{91.2} \\
\bottomrule
\end{tabular} }
\end{table*}

We evaluate \ours against previous state-of-the-art approaches under the standard 5-way K-shot setting. Tables~\ref{table:full_kinetics}, \ref{table:small_ucf_hmdb}, and \ref{table:finegym} present our results on SSV2 Full and Kinetics (K=1-5), SSV2 Small/UCF-101/HMDB-51 (K=1,3,5), and FineGym respectively.
On SSV2 Full, \ours consistently outperforms TATs~\cite{tats} with gains of 3.8\%, 2.8\%, 3.2\%, 5.3\%, and 2.1\% across 1-5 shots respectively. For Kinetics, we achieve improvements of 1.0\% and 1.2\% in 1-shot and 2-shot settings, with comparable performance in higher shots. The modest gains reflect Kinetics' inherent appearance bias, where actions are primarily distinguishable through static cues, reducing the effectiveness of our motion-focused contributions.
SSV2 Small demonstrates significant improvements with gains of 3.5\%, 5.3\%, and 4.5\% across 1,3,5-shot settings. UCF-101 shows consistent improvements of 2.0\%, 0.5\%, and 2.4\%, while HMDB-51 exhibits substantial gains of 9.8\%, 8.2\%, and 5.3\% across respective shots. FineGym follows this trend with improvements of 2.6\%, 2.3\%, and 2.0\% for 1,3,5-shot settings. 
Furthermore, when varying N-way settings (Table~\ref{table:nway}), \ours maintains its superior performance with consistent gains of 3-4\% on SSV2 Full and 1-2\% on Kinetics in the 1-shot setting. These substantial improvements across diverse datasets, shot settings, and N-way configurations demonstrate our approach's robust and superior nature.

\begin{figure}[t!]
    \centering
    \includegraphics[width=\columnwidth]{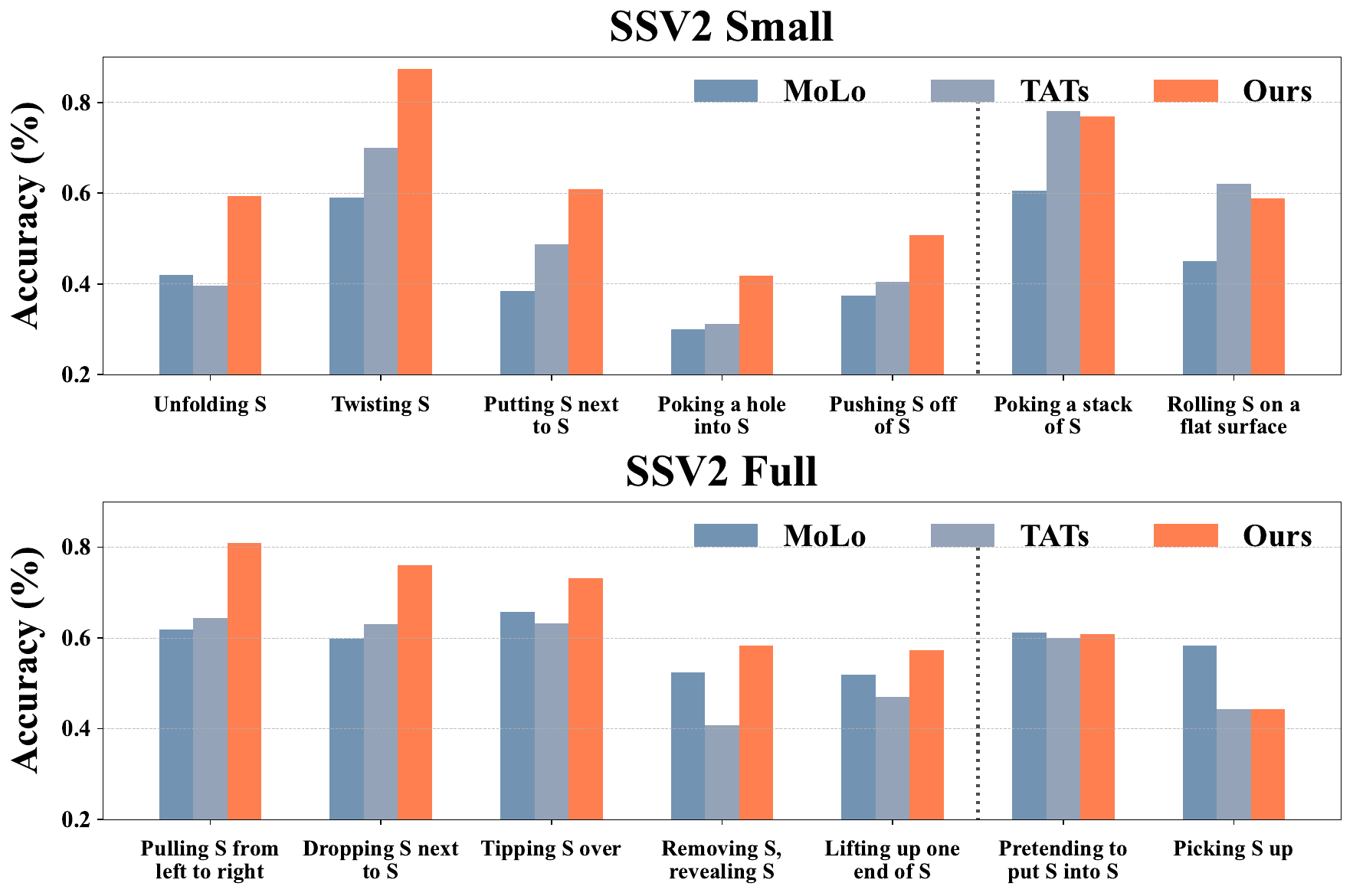}
    \caption{ Class-wise accuracy comparison between MoLo~\cite{molo}, TATs~\cite{tats} and our method. Left: classes where our approach shows performance gains. Right: classes without improvement. \vspace{-3em}}
    \label{fig:classwise}
\end{figure}

\paragraph{Class-wise performance anlaysis.} 
Figure~\ref{fig:classwise} presents a class-wise performance comparison between our method and previous approaches~\cite{molo,tats} on both SSV2 Small and SSV2 Full splits. Our method demonstrates consistent improvements across classes through the combined benefits of semantic-aware sampling and motion modules. The semantic sampling strategy ensures better tracking of smaller objects, while our motion modules capture complex temporal dynamics effectively. On SSV2 Small, classes like `Unfolding something' and `Twisting something' show notable improvements, particularly benefiting from our motion-aware architecture. Similarly, SSV2 Full exhibits enhanced performance in classes such as `Pulling something from left to right' and `Dropping something next to something'. However, our analysis also reveals limitations in handling rapid motions causing blur (e.g., `Rolling something on flat surface') and significant camera movements (e.g., `Picking something up'), where point tracking becomes challenging.

\begin{table*}[t!]
\addtolength{\tabcolsep}{2pt} %
\caption{Comparison of few-shot action accuracy (1, 3 and 5 shots) on SSV2 Small, UCF-101, and HMDB-51 datasets versus contemporary methods. Best results are bolded, second-best underlined, and ``-'' indicates unavailable data.}
\label{table:small_ucf_hmdb}
\resizebox{\textwidth}{!}{
\begin{tabular}{@{}llccc|ccc|ccc@{}}

\toprule
& & \multicolumn{3}{c}{\textbf{SSV2 Small}} & \multicolumn{3}{c}{\textbf{UCF-101}}  & \multicolumn{3}{c}{\textbf{HMDB-51}}  \\ \cmidrule(l){3-5} \cmidrule(l){6-8} \cmidrule(l){9-11} 
\textbf{Method} & \textbf{Reference} & \multicolumn{1}{l}{1-shot} & \multicolumn{1}{l}{3-shot} & \multicolumn{1}{l}{5-shot} & \multicolumn{1}{l}{1-shot} & \multicolumn{1}{l}{3-shot} & \multicolumn{1}{l}{5-shot} & \multicolumn{1}{l}{1-shot} & \multicolumn{1}{l}{3-shot} & \multicolumn{1}{l}{5-shot} \\
\midrule
OTAM~\cite{otam}& CVPR'20& 36.4& 45.9& 48.0& 79.9& 87.0& 88.9& 54.5& 65.7& 68.0 \\
TRX~\cite{trx}& CVPR'21& 36.0& 51.9& 56.7& 78.2& 92.4& 96.1& 53.1& 66.8& 75.6 \\
STRM~\cite{strm}& CVPR'22& 37.1& 49.2& 55.3& 80.5& 92.7& 96.9& 52.3& 67.4& 77.3 \\
MTFAN~\cite{mtfan}& CVPR'22& -& -& -& 84.8&- & 95.1&- & -& - \\
HYRSM~\cite{hyrsm}& CVPR'22& 40.6&52.3& 56.1& 83.9& 93.0& 94.7& 60.3& 71.7& 76.0 \\
HCL~\cite{hcl}& ECCV'22& 38.7& 49.1& 55.4& 82.5& 91.0& 93.9& 59.1& 71.2& 76.3 \\
Nguyen et al~\cite{nguyen2022inductive}& ECCV'22&- & -& -& -& -&- & 59.6& -& 76.9 \\
Huang et al~\cite{huang2022compound}& ECCV'22& 38.9& -& 61.6& 71.4& - & 91.0& 60.1& -& 77.0 \\
MoLo~\cite{molo}& CVPR'23& 42.7& 52.9& 56.4& 86.0& 93.5& 95.5& 60.8& 72.0& 77.4 \\
GgHM~\cite{gghm}& ICCV'23&- &- & -& 85.2&- & 96.3& 61.2&- & 76.9 \\
RFPL~\cite{rfpl}& ICCV'23& -& -&- & 82.5& 94.1& 96.3& -& -& - \\
CCLN~\cite{ccln}& PAMI'24&- &- &- & 86.9& 94.2& 96.1& 65.1& \underline{76.2}& 78.8 \\
HYRSM++~\cite{hyrsm++}& PR'24& 42.8& 52.4& 58.0& 85.8& 93.5& 95.9& 61.5& 72.7& 76.4 \\

TATs~\cite{tats} & ECCV'24 & \underline{47.9}& \underline{60.0}& \underline{64.4}& 92.0 & \underline{96.8}& 95.5& 60.0& 71.8& 77.0 \\
TEAM~\cite{Lee_2025_CVPR} & CVPR'25 & 47.2 & - & 63.1 & \textbf{94.5 }& - & \textbf{98.8} & \textbf{70.9} & - & \textbf{85.5} \\
\midrule
\textbf{\ours}& -& \textbf{53.4}& \textbf{65.3}& \textbf{68.9}& \underline{94.0}& \textbf{97.3}& \underline{97.9}& \underline{69.8}& \textbf{80.0}&\underline{ 82.3} \\
\bottomrule
\end{tabular}
}

\end{table*}

\begin{table*}[t]
\addtolength{\tabcolsep}{2pt} 
\caption{Comparative N-way 1-shot classification accuracy (N=5-10) on Kinetics and SSV2 Full datasets versus contemporary methods. Best and second-best results are bolded and underlined, respectively.}
\label{table:nway}
\resizebox{\textwidth}{!}{
\begin{tabular}{@{}lcccccc|cccccc@{}}

\toprule
& \multicolumn{6}{c}{\textbf{SSV2 Full}} & \multicolumn{6}{c}{\textbf{Kinetics}}  \\ 
\cmidrule(l){2-7} \cmidrule(l){8-13} 

\textbf{Method} &  \multicolumn{1}{l}{5-way} & \multicolumn{1}{l}{6-way} & \multicolumn{1}{l}{7-way} & \multicolumn{1}{l}{8-way} & \multicolumn{1}{l}{9-way} & \multicolumn{1}{l}{10-way}  &  \multicolumn{1}{l}{5-way} & \multicolumn{1}{l}{6-way} & \multicolumn{1}{l}{7-way} & \multicolumn{1}{l}{8-way} & \multicolumn{1}{l}{9-way} & \multicolumn{1}{l}{10-way} \\
\midrule
OTAM~\cite{otam} & 42.8 & 38.6 & 35.1 & 32.3 & 30.0 & 28.2 & 72.2 & 68.7 & 66.0 & 63.0 & 61.9 & 59.0 \\
TRX~\cite{trx} & 42.0 & 41.5 & 36.1 & 33.6 & 32.0 & 30.3 & 63.6 & 59.4 & 56.7 & 54.6 & 53.2 & 51.1 \\
HyRSM~\cite{hyrsm} & 54.3 & 50.1 & 45.8 & 44.3 & 42.1 & 40.0 & 73.7 & 69.5 & 66.6 & 65.5 & 63.4 & 61.0 \\
MoLo~\cite{molo} & 56.6 & 51.6 & 48.1 & 44.8 & 42.5 & 40.3 & 74.0 & 69.7 & 67.4 & 65.8 & 63.5 & 61.3 \\
TATs~\cite{tats} & \underline{57.7} & \underline{55.7} & \underline{52.5} & \underline{50.0} & \underline{47.0} & \underline{45.8} & \underline{81.9} & \underline{79.0} & \underline{76.1} & \underline{75.2} & \underline{72.2} & \underline{72.0} \\
\midrule
\textbf{\ours} & \textbf{61.5} & \textbf{59.1} & \textbf{56.5} & \textbf{54.6} & \textbf{51.4} & \textbf{49.1} & \textbf{82.9} & \textbf{80.2} & \textbf{78.5} & \textbf{76.8} & \textbf{75.5} & \textbf{73.3} \\
\bottomrule
\end{tabular}}

\end{table*}

\subsection{Ablation Analysis}
We conduct an ablation study to evaluate key design decisions. Through systematic experimentation, we analyze the impact of each component on model performance and provide empirical justification for our final configuration. We also show additional ablations in our supplementary.

\paragraph{Impact of each component.} 
\begin{table}[t]
\addtolength{\tabcolsep}{1pt} 
\caption{Impact of each component on Trokens, demonstrating the relative contribution of individual elements with the final row representing our final setting.}
\label{table:per_component}
\resizebox{\columnwidth}{!}{
\begin{tabular}{@{}ccccccc@{}}
\toprule 
\multirow{2}{*}{\textbf{\begin{tabular}[c]{@{}c@{}}Semantic-aware\\sampling\end{tabular}}} & 
\multirow{2}{*}{\textbf{\begin{tabular}[c]{@{}c@{}}Intra\\motion\end{tabular}}} & 
\multirow{2}{*}{\textbf{\begin{tabular}[c]{@{}c@{}}Inter\\motion\end{tabular}}} & 
\multicolumn{2}{c}{\textbf{SSV2 Small}} & \multicolumn{2}{c}{\textbf{SSV2 Full}} \\
\cmidrule(l){4-5}  \cmidrule(l){6-7} 
 &  &  & \multicolumn{1}{l}{1-shot} & \multicolumn{1}{l}{5-shot} & \multicolumn{1}{l}{1-shot} & \multicolumn{1}{l}{5-shot}\\
\midrule
\cross & \cross & \cross & 47.9 & 64.4 & 57.7 & 74.6 \\
\tick & \cross & \cross & 49.9 & 65.7 & 58.6 & 74.2 \\
\tick & \tick & \cross & 51.8 & 67.7 & 61.3 & 75.7 \\
\tick & \cross & \tick & 52.2 & 67.3 & 60.6 & 74.8 \\
\tick & \tick & \tick & \textbf{53.4} & \textbf{68.9} & \textbf{61.5} & \textbf{76.7 }\\
 \bottomrule
\end{tabular}}
\end{table}

Table~\ref{table:per_component} presents an ablation study of our key components. Starting from the baseline configuration~\cite{tats}, we first evaluate our semantic-aware point sampling strategy, which yields improvements of 2\% and 1.3\% on SSV2 Small (1-shot and 5-shot), and 0.9\% on SSV2 Full (1-shot). Given these gains, we retain this sampling strategy for subsequent experiments. We then analyze our motion modules independently. The intra-motion module alone achieves gains of $\sim$2\% on both shots of SSV2 Small and up to 3.3\% on SSV2 Full. Similarly, the inter-motion module independently shows improvements of 1.3--1.6\% on SSV2 Small and up to 2\% on SSV2 Full. When combined, these modules yield consistent improvements of 1--2\% across all settings, suggesting they capture complementary motion information that enhances performance.

\paragraph{Analysis of Intra-motion module.} 
\begin{table*}[!t]
\begin{minipage}[t]{0.237\linewidth}
\addtolength{\tabcolsep}{5pt} %
\caption{Comparison of few-shot action accuracy (1, 3 and 5 shots) on the FineGym dataset.}
\label{table:finegym}
\centering
\resizebox{1.0\columnwidth}{!}{
\begin{tabular}{@{}lccc@{}}
\toprule
&  \multicolumn{3}{c}{\textbf{FineGym}}  \\
\cmidrule{2-4}
\textbf{Method} &  1-shot & 3-shot  & 5-shot  \\
\midrule
MoLo~\cite{molo} &  73.3 & 80.2 &84.8 \\
TATs~\cite{tats} & \underline{81.8} &\underline{86.0} &\underline{87.9} \\
\midrule
\textbf{\ours} &  \textbf{84.4 }& \textbf{88.3 } & \textbf{89.9} \\
\bottomrule
\end{tabular}}
\end{minipage}
\hfill
\begin{minipage}[t]{0.347\linewidth}
\addtolength{\tabcolsep}{1pt} 
\caption{Comparative analysis of intra-motion module variants independent of inter-motion module. }
\label{table:intra_motion}
\resizebox{\columnwidth}{!}{
\begin{tabular}{@{}lccc@{}}
\toprule 
& \multicolumn{3}{c}{\textbf{SSV2 Full}} \\
  \cmidrule(l){2-4} 
\textbf{Intra-motion module }& \multicolumn{1}{l}{1-shot} & \multicolumn{1}{l}{3-shot} & \multicolumn{1}{l}{5-shot} \\
\midrule
Displacement only & 59.9 & 71.6 & 74.2 \\
HoD (ours) & \textbf{61.3} & \textbf{73.6} & \textbf{75.7} \\
 \bottomrule
\end{tabular}}
\end{minipage}
\hfill
\begin{minipage}[t]{0.347\linewidth}
\addtolength{\tabcolsep}{1pt} 
\renewcommand{\arraystretch}{0.925}
\caption{Trainable parameter analysis. * represents our implementation of \cite{tats} with increased parameters for fair comparison.}
\label{table:params}
\resizebox{\columnwidth}{!}{
\begin{tabular}{@{}lccccc@{}}
\toprule 
& & \multicolumn{2}{c}{\textbf{SSV2 Small}} & \multicolumn{2}{c}{\textbf{SSV2 Full}} \\
\cmidrule(l){3-4}  \cmidrule(l){5-6} 
\textbf{Method} & \textbf{Params} & \multicolumn{1}{l}{1-shot} & \multicolumn{1}{l}{5-shot} & \multicolumn{1}{l}{1-shot} & \multicolumn{1}{l}{5-shot} \\
\midrule
TATs~\cite{tats} & 11.8 M & 47.9 & 64.4 & 57.7 & 74.6 \\
TATs*~\cite{tats} & 18.1 M & 48.0 & 63.0 & 59.6 & 74.3 \\
\midrule
\textbf{\ours} & 17.4 M & \textbf{53.4} & \textbf{68.9} & \textbf{61.5 }& \textbf{76.7} \\
 \bottomrule
\end{tabular}}
\end{minipage}
\end{table*}

Table~\ref{table:intra_motion} compares our choice of Histogram of Oriented Displacement (HoD) features against displacement-only features for intra-motion representation. To isolate the impact of feature choice, we conduct all experiments with the inter-motion module disabled. In this controlled setting, HoD features consistently outperform displacement-only features on SSV2 Full, yielding gains of 2.4\%, 2.0\%, and 1.5\% for 1-shot, 3-shot, and 5-shot settings, respectively. These results demonstrate that incorporating directional information through HoD features captures richer trajectory patterns compared to using displacement information alone.

\begin{figure}[t]
    \centering
    \includegraphics[width=1.0\linewidth]{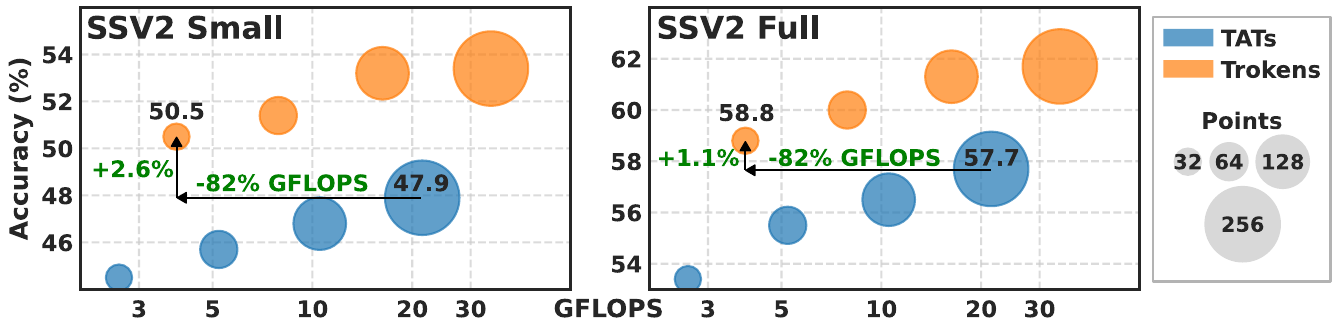}
    \caption{\textbf{FLOPS \vs Accuracy Comparison} for Trokens' semantically-guided sampling against uniform sampling (TATs) over serveral point counts. Trokens' sampling enables higher performance at lower sampling rates boosting overall efficience.}
    \label{fig:flops}
\end{figure}

\paragraph{Performance \vs Efficiency Analysis.} In Fig.~\ref{fig:flops}, we present analysis including runtime FLOPs to  demonstrate the efficiency benefits of our method.
\ours offsets the computational cost of clustering by efficiently selecting points that achieve higher performance with fewer tracking points.
For both SSV2 Small and SSV2 Full, our method with just 32 points surpasses TATs (uniform sampling) with 256 points, while using 82\% fewer inference-time FLOPs overall. Note that FLOP calculations exclude the DINO feature extractor as it remains constant across methods.

\paragraph{Effect of trainable parameters.} To address potential concerns about performance gains stemming from increased model capacity, we implemented a modified version of~\cite{tats} with expanded parameters to match our model size. We maintaining the rest of their original configuration. Table~\ref{table:params} shows that this parameter-matched re-implementation of~\cite{tats} does not yield improved performance. This demonstrates that our performance gains stem from the effectiveness of our motion modules in capturing complementary information, rather than from increased model capacity.

\subsection{Qualitative Analysis}
\begin{figure}[t!]
    \centering
    \includegraphics[width=\columnwidth]{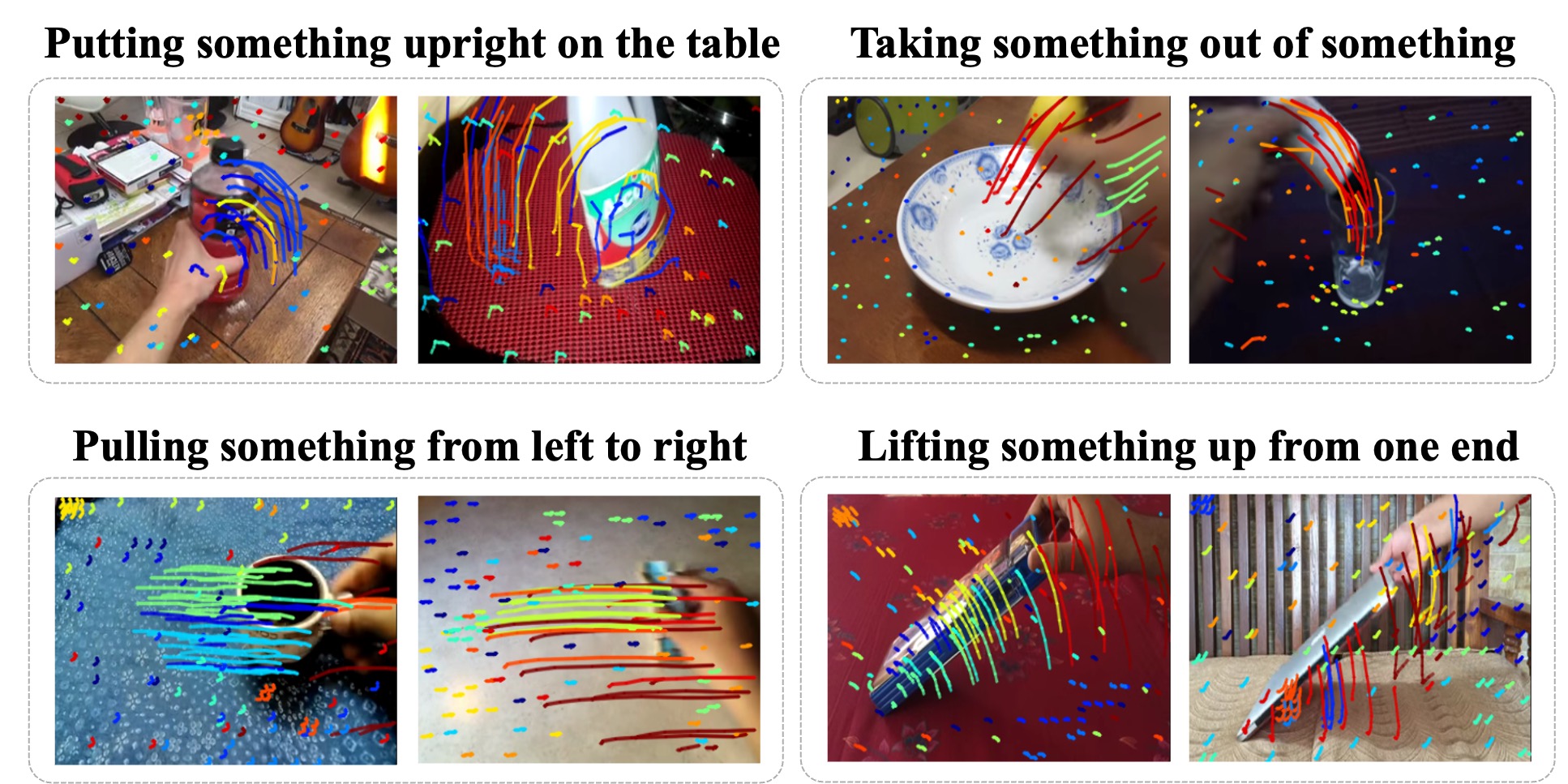}
    \caption{Visualization of action trajectory similarities across four classes, where semantic-based sampling enables object-focused trajectories. Each quadrant demonstrates intra-class motion consistency while maintaining inter-class discriminative features. }
    \label{fig:qual}
\end{figure}

Figure~\ref{fig:qual} demonstrates our semantic-aware point sampling through trajectory visualization. Our method concentrates tracking points on action-relevant objects for meaningful motion capture. The top-right quadrant showing ``Taking something out of something'' tracks both a lemon and bottle, capturing the essential lifting motion. Trajectories across different examples of the same action show striking similarities while remaining distinct from other actions. This intra-class similarity and inter-class variation shows how our semantic sampling enhances action recognition by focusing on meaningful motion patterns.

\section{Limitations and Future Work}

While our approach demonstrates strong performance overall, it faces certain limitations. On datasets like Kinetics, our motion-focused approach provides limited performance gains as its actions are distinguishable from appearance alone. Additionally, our method faces challenges inherent to point tracking: it becomes vulnerable to rapid motions that cause motion blur. Significant camera movement can disrupt point trajectory consistency, impacting performance in classes involving substantial viewpoint changes. These limitations primarily stem from the fundamental challenges in point tracking rather than our architectural choices, suggesting potential future directions in developing more robust tracking mechanisms. Future work can further explore recent works in 3D point tracking~\cite{koppula2024tapvid,SpatialTracker,xiao2025spatialtrackerv23dpointtracking,zhang2025tapip3d,vggt,feng2025st4rtrack} and improve upon current tracking robustness to address these fundamental challenges. While we demonstrate the effectiveness of our approach in few-shot action recognition, we hope \ours inspires  research in both full classification settings and broader video understanding tasks.

\section{Conclusion}
We introduced \ours, a novel approach for few-shot action recognition using semantic-aware motion trajectory tokens. Our method addresses key challenges in point tracking-based video understanding through semantic-aware sampling and relational motion modeling that captures intra-trajectory dynamics and inter-trajectory relationships. State-of-the-art performance across six challenging benchmarks validates our core insight that combining semantic guidance with explicit motion modeling provides a robust foundation for understanding human actions in limited-data scenarios while opening promising directions for future research in dynamic scene understanding.

\section*{Acknowledgments} 
We would like to thank Matthew Gwilliam for suggesting the name \emph{Trokens} and for his valuable feedback for the manuscript. This work was partially supported by NSF CAREER Award (\#2238769) to AS. The authors acknowledge UMD’s supercomputing resources made available for conducting this research. The U.S. Government is authorized to reproduce and distribute reprints for Governmental purposes notwithstanding any copyright annotation thereon. The views and conclusions contained herein are those of the authors and should not be interpreted as necessarily representing the official policies or endorsements, either expressed or implied, of NSF or the U.S. Government.

{
    \small
    \bibliographystyle{ieeenat_fullname}
    \bibliography{main}

\begin{thebibliography}{100}
\providecommand{\natexlab}[1]{#1}
\providecommand{\url}[1]{\texttt{#1}}
\expandafter\ifx\csname urlstyle\endcsname\relax
  \providecommand{\doi}[1]{doi: #1}\else
  \providecommand{\doi}{doi: \begingroup \urlstyle{rm}\Url}\fi

\bibitem[Abdul-Azim and Hemayed(2015)]{abdul2015human}
Haiam~A Abdul-Azim and Elsayed~E Hemayed.
\newblock Human action recognition using trajectory-based representation.
\newblock \emph{Egyptian Informatics Journal}, 16\penalty0 (2):\penalty0 187--198, 2015.

\bibitem[Arnab et~al.(2021)Arnab, Dehghani, Heigold, Sun, Lu{\v{c}}i{\'c}, and Schmid]{arnab2021vivit}
Anurag Arnab, Mostafa Dehghani, Georg Heigold, Chen Sun, Mario Lu{\v{c}}i{\'c}, and Cordelia Schmid.
\newblock Vivit: A video vision transformer.
\newblock In \emph{Proceedings of the IEEE/CVF international conference on computer vision}, pages 6836--6846, 2021.

\bibitem[Bertasius et~al.(2021)Bertasius, Wang, and Torresani]{timesformer}
Gedas Bertasius, Heng Wang, and Lorenzo Torresani.
\newblock Is space-time attention all you need for video understanding?
\newblock In \emph{Proceedings of the International Conference on Machine Learning (ICML)}, 2021.

\bibitem[Bishay et~al.(2019)Bishay, Zoumpourlis, and Patras]{bishay2019tarn}
Mina Bishay, Georgios Zoumpourlis, and Ioannis Patras.
\newblock Tarn: Temporal attentive relation network for few-shot and zero-shot action recognition.
\newblock \emph{arXiv preprint arXiv:1907.09021}, 2019.

\bibitem[Bolya et~al.(2022)Bolya, Fu, Dai, Zhang, Feichtenhofer, and Hoffman]{bolya2022token}
Daniel Bolya, Cheng-Yang Fu, Xiaoliang Dai, Peizhao Zhang, Christoph Feichtenhofer, and Judy Hoffman.
\newblock Token merging: Your vit but faster.
\newblock \emph{arXiv preprint arXiv:2210.09461}, 2022.

\bibitem[Cao et~al.(2024{\natexlab{a}})Cao, Zhang, Lv, Min, and Zhang]{cao2024exploring}
Congqi Cao, Yueran Zhang, Qinyi Lv, Lingtong Min, and Yanning Zhang.
\newblock Exploring the adaptation strategy of clip for few-shot action recognition.
\newblock In \emph{Proceedings of the 1st International Workshop on Efficient Multimedia Computing under Limited}, pages 39--48, 2024{\natexlab{a}}.

\bibitem[Cao et~al.(2024{\natexlab{b}})Cao, Zhang, Yu, Lv, Min, and Zhang]{cao2024task}
Congqi Cao, Yueran Zhang, Yating Yu, Qinyi Lv, Lingtong Min, and Yanning Zhang.
\newblock Task-adapter: Task-specific adaptation of image models for few-shot action recognition.
\newblock In \emph{Proceedings of the 32nd ACM International Conference on Multimedia}, pages 9038--9047, 2024{\natexlab{b}}.

\bibitem[Cao et~al.(2020)Cao, Ji, Cao, Chang, and Niebles]{otam}
Kaidi Cao, Jingwei Ji, Zhangjie Cao, Chien-Yi Chang, and Juan~Carlos Niebles.
\newblock Few-shot video classification via temporal alignment.
\newblock In \emph{Proceedings of the IEEE/CVF Conference on Computer Vision and Pattern Recognition}, pages 10618--10627, 2020.

\bibitem[Carreira and Zisserman(2017)]{kinetics}
Jo{\~a}o Carreira and Andrew Zisserman.
\newblock Quo vadis, action recognition? a new model and the kinetics dataset.
\newblock \emph{2017 IEEE Conference on Computer Vision and Pattern Recognition (CVPR)}, pages 4724--4733, 2017.

\bibitem[Cho et~al.(2024)Cho, Huang, Nam, An, Kim, and Lee]{locotrack}
Seokju Cho, Jiahui Huang, Jisu Nam, Honggyu An, Seungryong Kim, and Joon-Young Lee.
\newblock Local all-pair correspondence for point tracking.
\newblock In \emph{European Conference on Computer Vision}, pages 306--325. Springer, 2024.

\bibitem[Dalal and Triggs(2005)]{dalal2005histograms}
Navneet Dalal and Bill Triggs.
\newblock Histograms of oriented gradients for human detection.
\newblock In \emph{2005 IEEE computer society conference on computer vision and pattern recognition (CVPR'05)}, pages 886--893. Ieee, 2005.

\bibitem[Deng et~al.(2024)Deng, Zhong, Li, Fu, Jiang, Ningbo, Qi, Xin, and Lam]{deng2024text}
Fuqin Deng, Jiaming Zhong, Nannan Li, Lanhui Fu, Bingchun Jiang, Yi Ningbo, Feng Qi, He Xin, and Tin~Lun Lam.
\newblock Text-guided graph temporal modeling for few-shot video classification.
\newblock \emph{Engineering Applications of Artificial Intelligence}, 137:\penalty0 109076, 2024.

\bibitem[Doersch et~al.(2022)Doersch, Gupta, Markeeva, Recasens, Smaira, Aytar, Carreira, Zisserman, and Yang]{doersch2022tap}
Carl Doersch, Ankush Gupta, Larisa Markeeva, Adri{\`a} Recasens, Lucas Smaira, Yusuf Aytar, Jo{\~a}o Carreira, Andrew Zisserman, and Yi Yang.
\newblock Tap-vid: A benchmark for tracking any point in a video.
\newblock \emph{Advances in Neural Information Processing Systems}, 35:\penalty0 13610--13626, 2022.

\bibitem[Doersch et~al.(2023)Doersch, Yang, Vecerik, Gokay, Gupta, Aytar, Carreira, and Zisserman]{doersch2023tapir}
Carl Doersch, Yi Yang, Mel Vecerik, Dilara Gokay, Ankush Gupta, Yusuf Aytar, Joao Carreira, and Andrew Zisserman.
\newblock Tapir: Tracking any point with per-frame initialization and temporal refinement.
\newblock \emph{arXiv preprint arXiv:2306.08637}, 2023.

\bibitem[Feng et~al.(2025)Feng, Zhang, Wang, Ye, Yu, Black, Darrell, and Kanazawa]{feng2025st4rtrack}
Haiwen Feng, Junyi Zhang, Qianqian Wang, Yufei Ye, Pengcheng Yu, Michael~J Black, Trevor Darrell, and Angjoo Kanazawa.
\newblock St4rtrack: Simultaneous 4d reconstruction and tracking in the world.
\newblock \emph{arXiv preprint arXiv:2504.13152}, 2025.

\bibitem[Fu et~al.(2020)Fu, Zhang, Wang, Fu, and Jiang]{fu2020depth}
Yuqian Fu, Li Zhang, Junke Wang, Yanwei Fu, and Yu-Gang Jiang.
\newblock Depth guided adaptive meta-fusion network for few-shot video recognition.
\newblock In \emph{Proceedings of the 28th ACM International Conference on Multimedia}, pages 1142--1151, 2020.

\bibitem[Gowayyed et~al.(2013{\natexlab{a}})Gowayyed, Torki, Hussein, and El-Saban]{gowayyed2013histogram}
Mohammad~Abdelaziz Gowayyed, Marwan Torki, Mohamed~Elsayed Hussein, and Motaz El-Saban.
\newblock Histogram of oriented displacements (hod): Describing trajectories of human joints for action recognition.
\newblock In \emph{IJCAI}, pages 1351--1357, 2013{\natexlab{a}}.

\bibitem[Gowayyed et~al.(2013{\natexlab{b}})Gowayyed, Torki, Hussein, and El-Saban]{hod}
Mohammad~Abdelaziz Gowayyed, Marwan Torki, Mohammed~Elsayed Hussein, and Motaz El-Saban.
\newblock Histogram of oriented displacements (hod): Describing trajectories of human joints for action recognition.
\newblock In \emph{Twenty-third international joint conference on artificial intelligence}, 2013{\natexlab{b}}.

\bibitem[Goyal et~al.(2017)Goyal, Kahou, Michalski, Materzynska, Westphal, Kim, Haenel, Fr{\"u}nd, Yianilos, Mueller-Freitag, Hoppe, Thurau, Bax, and Memisevic]{ssv2}
Raghav Goyal, Samira~Ebrahimi Kahou, Vincent Michalski, Joanna Materzynska, Susanne Westphal, Heuna Kim, Valentin Haenel, Ingo Fr{\"u}nd, Peter~N. Yianilos, Moritz Mueller-Freitag, Florian Hoppe, Christian Thurau, Ingo Bax, and Roland Memisevic.
\newblock The “something something” video database for learning and evaluating visual common sense.
\newblock \emph{2017 IEEE International Conference on Computer Vision (ICCV)}, pages 5843--5851, 2017.

\bibitem[Guo et~al.(2024)Guo, Wang, Qi, Jin, Zhu, and Sun]{guo2024multi}
Fei Guo, YiKang Wang, Han Qi, Wenping Jin, Li Zhu, and Jing Sun.
\newblock Multi-view distillation based on multi-modal fusion for few-shot action recognition (clip-mdmf).
\newblock \emph{Knowledge-Based Systems}, 304:\penalty0 112539, 2024.

\bibitem[Hamilton et~al.(2022)Hamilton, Zhang, Hariharan, Snavely, and Freeman]{hamilton2022unsupervised}
Mark Hamilton, Zhoutong Zhang, Bharath Hariharan, Noah Snavely, and William~T. Freeman.
\newblock Unsupervised semantic segmentation by distilling feature correspondences.
\newblock In \emph{International Conference on Learning Representations}, 2022.

\bibitem[Harley et~al.(2022)Harley, Fang, and Fragkiadaki]{harley2022particle}
Adam~W Harley, Zhaoyuan Fang, and Katerina Fragkiadaki.
\newblock Particle video revisited: Tracking through occlusions using point trajectories.
\newblock In \emph{European Conference on Computer Vision}, pages 59--75. Springer, 2022.

\bibitem[Harley et~al.(2025)Harley, You, Sun, Zheng, Raghuraman, Gu, Liang, Chu, Dave, Tokmakov, et~al.]{alltracker}
Adam~W Harley, Yang You, Xinglong Sun, Yang Zheng, Nikhil Raghuraman, Yunqi Gu, Sheldon Liang, Wen-Hsuan Chu, Achal Dave, Pavel Tokmakov, et~al.
\newblock Alltracker: Efficient dense point tracking at high resolution.
\newblock \emph{arXiv preprint arXiv:2506.07310}, 2025.

\bibitem[He et~al.(2023)He, Yang, Wang, Wu, Chen, Huang, Ren, Lim, and Shrivastava]{he2023towards}
Bo He, Xitong Yang, Hanyu Wang, Zuxuan Wu, Hao Chen, Shuaiyi Huang, Yixuan Ren, Ser-Nam Lim, and Abhinav Shrivastava.
\newblock Towards scalable neural representation for diverse videos.
\newblock In \emph{Proceedings of the IEEE/CVF Conference on Computer Vision and Pattern Recognition}, pages 6132--6142, 2023.

\bibitem[Huang et~al.(2019)Huang, Wang, Zhang, Yan, and He]{huang2019dynamic}
Shuaiyi Huang, Qiuyue Wang, Songyang Zhang, Shipeng Yan, and Xuming He.
\newblock Dynamic context correspondence network for semantic alignment.
\newblock In \emph{Proceedings of the IEEE/CVF International Conference on Computer Vision}, pages 2010--2019, 2019.

\bibitem[Huang et~al.(2020)Huang, Wang, and He]{huang2020confidence}
Shuaiyi Huang, Qiuyue Wang, and Xuming He.
\newblock Confidence-aware adversarial learning for self-supervised semantic matching.
\newblock In \emph{Chinese Conference on Pattern Recognition and Computer Vision (PRCV)}, pages 91--103. Springer, 2020.

\bibitem[Huang et~al.(2022{\natexlab{a}})Huang, Yang, He, Zhang, He, and Shrivastava]{huang2022learning}
Shuaiyi Huang, Luyu Yang, Bo He, Songyang Zhang, Xuming He, and Abhinav Shrivastava.
\newblock Learning semantic correspondence with sparse annotations.
\newblock In \emph{European Conference on Computer Vision}, pages 267--284. Springer, 2022{\natexlab{a}}.

\bibitem[Huang et~al.(2024{\natexlab{a}})Huang, Huang, Yu, Lan, Radhakrishnan, Alvarez, Shrivastava, and Anandkumar]{huang2024point}
Shuaiyi Huang, De-An Huang, Zhiding Yu, Shiyi Lan, Subhashree Radhakrishnan, Jose~M Alvarez, Abhinav Shrivastava, and Anima Anandkumar.
\newblock What is point supervision worth in video instance segmentation?
\newblock In \emph{Proceedings of the IEEE/CVF Conference on Computer Vision and Pattern Recognition Workshops (CVPRW)}, pages 2671--2681, 2024{\natexlab{a}}.

\bibitem[Huang et~al.(2024{\natexlab{b}})Huang, Levy, Jiang, Anandkumar, Zhu, Fan, Huang, and Shrivastava]{huang2024ardup}
Shuaiyi Huang, Mara Levy, Zhenyu Jiang, Anima Anandkumar, Yuke Zhu, Linxi Fan, De-An Huang, and Abhinav Shrivastava.
\newblock Ardup: Active region video diffusion for universal policies.
\newblock \emph{arXiv preprint arXiv:2406.13301}, 2024{\natexlab{b}}.

\bibitem[Huang et~al.(2024{\natexlab{c}})Huang, Suri, Gupta, Rambhatla, Lim, and Shrivastava]{huang2024uvis}
Shuaiyi Huang, Saksham Suri, Kamal Gupta, Sai~Saketh Rambhatla, Ser-nam Lim, and Abhinav Shrivastava.
\newblock Uvis: Unsupervised video instance segmentation.
\newblock In \emph{Proceedings of the IEEE/CVF Conference on Computer Vision and Pattern Recognition Workshops (CVPRW)}, pages 2682--2692, 2024{\natexlab{c}}.

\bibitem[Huang et~al.(2025)Huang, Levy, Gupta, Ekpo, Zheng, and Shrivastava]{huang2025trend}
Shuaiyi Huang, Mara Levy, Anubhav Gupta, Daniel Ekpo, Ruijie Zheng, and Abhinav Shrivastava.
\newblock Trend: Tri-teaching for robust preference-based reinforcement learning with demonstrations.
\newblock \emph{arXiv preprint arXiv:2505.06079}, 2025.

\bibitem[Huang et~al.(2024{\natexlab{d}})Huang, Zhang, Li, Zhang, Wang, Dong, Jin, Ogawa, and Haseyama]{manta}
Wenbo Huang, Jinghui Zhang, Guang Li, Lei Zhang, Shuoyuan Wang, Fang Dong, Jiahui Jin, Takahiro Ogawa, and Miki Haseyama.
\newblock Manta: Enhancing mamba for few-shot action recognition of long sub-sequence.
\newblock \emph{arXiv preprint arXiv:2412.07481}, 2024{\natexlab{d}}.

\bibitem[Huang et~al.(2022{\natexlab{b}})Huang, Yang, and Sato]{huang2022compound}
Yifei Huang, Lijin Yang, and Yoichi Sato.
\newblock Compound prototype matching for few-shot action recognition.
\newblock In \emph{European Conference on Computer Vision}, pages 351--368. Springer, 2022{\natexlab{b}}.

\bibitem[Karaev et~al.(2024{\natexlab{a}})Karaev, Makarov, Wang, Neverova, Vedaldi, and Rupprecht]{cotracker3}
Nikita Karaev, Iurii Makarov, Jianyuan Wang, Natalia Neverova, Andrea Vedaldi, and Christian Rupprecht.
\newblock Cotracker3: Simpler and better point tracking by pseudo-labelling real videos.
\newblock \emph{arXiv preprint arXiv:2410.11831}, 2024{\natexlab{a}}.

\bibitem[Karaev et~al.(2024{\natexlab{b}})Karaev, Rocco, Graham, Neverova, Vedaldi, and Rupprecht]{cotracker}
Nikita Karaev, Ignacio Rocco, Benjamin Graham, Natalia Neverova, Andrea Vedaldi, and Christian Rupprecht.
\newblock Cotracker: It is better to track together.
\newblock In \emph{European Conference on Computer Vision}, pages 18--35. Springer, 2024{\natexlab{b}}.

\bibitem[Kliper-Gross et~al.(2011)Kliper-Gross, Hassner, and Wolf]{kliper2011one}
Orit Kliper-Gross, Tal Hassner, and Lior Wolf.
\newblock One shot similarity metric learning for action recognition.
\newblock In \emph{Similarity-Based Pattern Recognition: First International Workshop, SIMBAD 2011, Venice, Italy, September 28-30, 2011. Proceedings 1}, pages 31--45. Springer, 2011.

\bibitem[Koppula et~al.(2024)Koppula, Rocco, Yang, Heyward, Carreira, Zisserman, Brostow, and Doersch]{koppula2024tapvid}
Skanda Koppula, Ignacio Rocco, Yi Yang, Joe Heyward, Jo{\~a}o Carreira, Andrew Zisserman, Gabriel Brostow, and Carl Doersch.
\newblock Tapvid-3d: A benchmark for tracking any point in 3d.
\newblock \emph{arXiv preprint arXiv:2407.05921}, 2024.

\bibitem[Kuehne et~al.(2011)Kuehne, Jhuang, Garrote, Poggio, and Serre]{hmdb}
Hilde Kuehne, Hueihan Jhuang, Est{\'i}baliz Garrote, Tomaso~A. Poggio, and Thomas Serre.
\newblock Hmdb: A large video database for human motion recognition.
\newblock \emph{2011 International Conference on Computer Vision}, pages 2556--2563, 2011.

\bibitem[Kumar et~al.(2024)Kumar, Padmanabhan, Luo, Rambhatla, and Shrivastava]{tats}
Pulkit Kumar, Namitha Padmanabhan, Luke Luo, Sai~Saketh Rambhatla, and Abhinav Shrivastava.
\newblock Trajectory-aligned space-time tokens for few-shot action recognition.
\newblock In \emph{European Conference on Computer Vision}, pages 474--493. Springer, 2024.

\bibitem[Kwon et~al.(2020)Kwon, Kim, Kwak, and Cho]{kwon2020motionsqueeze}
Heeseung Kwon, Manjin Kim, Suha Kwak, and Minsu Cho.
\newblock Motionsqueeze: Neural motion feature learning for video understanding.
\newblock In \emph{Computer Vision--ECCV 2020: 16th European Conference, Glasgow, UK, August 23--28, 2020, Proceedings, Part XVI 16}, pages 345--362. Springer, 2020.

\bibitem[Laptev et~al.(2008)Laptev, Marszalek, Schmid, and Rozenfeld]{laptev2008learning}
Ivan Laptev, Marcin Marszalek, Cordelia Schmid, and Benjamin Rozenfeld.
\newblock Learning realistic human actions from movies.
\newblock In \emph{2008 IEEE conference on computer vision and pattern recognition}, pages 1--8. IEEE, 2008.

\bibitem[Lee et~al.(2025)Lee, Moon, Seong, and Heo]{Lee_2025_CVPR}
SuBeen Lee, WonJun Moon, Hyun~Seok Seong, and Jae-Pil Heo.
\newblock Temporal alignment-free video matching for few-shot action recognition.
\newblock In \emph{Proceedings of the Computer Vision and Pattern Recognition Conference (CVPR)}, pages 5412--5421, 2025.

\bibitem[Li et~al.(2024)Li, Liu, Wang, and Yu]{li2024frame}
Bozheng Li, Mushui Liu, Gaoang Wang, and Yunlong Yu.
\newblock Frame order matters: A temporal sequence-aware model for few-shot action recognition.
\newblock \emph{arXiv preprint arXiv:2408.12475}, 2024.

\bibitem[Moing et~al.(2023)Moing, Ponce, and Schmid]{moing2023dense}
Guillaume~Le Moing, Jean Ponce, and Cordelia Schmid.
\newblock Dense optical tracking: Connecting the dots.
\newblock \emph{arXiv preprint arXiv:2312.00786}, 2023.

\bibitem[Nguyen et~al.(2022)Nguyen, Tran, Nguyen, Hua, and Nguyen]{nguyen2022inductive}
Khoi~D Nguyen, Quoc-Huy Tran, Khoi Nguyen, Binh-Son Hua, and Rang Nguyen.
\newblock Inductive and transductive few-shot video classification via appearance and temporal alignments.
\newblock In \emph{European Conference on Computer Vision}, pages 471--487. Springer, 2022.

\bibitem[Ni et~al.(2022)Ni, Liu, Wen, Ji, Xiao, and Yang]{ni2022multimodal}
Xinzhe Ni, Yong Liu, Hao Wen, Yatai Ji, Jing Xiao, and Yujiu Yang.
\newblock Multimodal prototype-enhanced network for few-shot action recognition.
\newblock \emph{arXiv preprint arXiv:2212.04873}, 2022.

\bibitem[Oquab et~al.(2023)Oquab, Darcet, Moutakanni, Vo, Szafraniec, Khalidov, Fernandez, Haziza, Massa, El-Nouby, Assran, Ballas, Galuba, Howes, Huang, Li, Misra, Rabbat, Sharma, Synnaeve, Xu, J{\'e}gou, Mairal, Labatut, Joulin, and Bojanowski]{dinov2}
Maxime Oquab, Timoth'ee Darcet, Th{\'e}o Moutakanni, Huy~Q. Vo, Marc Szafraniec, Vasil Khalidov, Pierre Fernandez, Daniel Haziza, Francisco Massa, Alaaeldin El-Nouby, Mahmoud Assran, Nicolas Ballas, Wojciech Galuba, Russ Howes, Po-Yao~(Bernie) Huang, Shang-Wen Li, Ishan Misra, Michael~G. Rabbat, Vasu Sharma, Gabriel Synnaeve, Huijiao Xu, Herv{\'e} J{\'e}gou, Julien Mairal, Patrick Labatut, Armand Joulin, and Piotr Bojanowski.
\newblock Dinov2: Learning robust visual features without supervision.
\newblock \emph{ArXiv}, abs/2304.07193, 2023.

\bibitem[Patrick et~al.(2021)Patrick, Campbell, Asano, Misra, Metze, Feichtenhofer, Vedaldi, and Henriques]{patrick2021keeping}
Mandela Patrick, Dylan Campbell, Yuki Asano, Ishan Misra, Florian Metze, Christoph Feichtenhofer, Andrea Vedaldi, and Joao~F Henriques.
\newblock Keeping your eye on the ball: Trajectory attention in video transformers.
\newblock \emph{Advances in neural information processing systems}, 34:\penalty0 12493--12506, 2021.

\bibitem[Perrett et~al.(2021)Perrett, Masullo, Burghardt, Mirmehdi, and Damen]{trx}
Toby Perrett, Alessandro Masullo, Tilo Burghardt, Majid Mirmehdi, and Dima Damen.
\newblock Temporal-relational crosstransformers for few-shot action recognition.
\newblock In \emph{Proceedings of the IEEE/CVF conference on computer vision and pattern recognition}, pages 475--484, 2021.

\bibitem[Per{\v{s}} et~al.(2010)Per{\v{s}}, Suli{\'c}, Kristan, Per{\v{s}}e, Polanec, and Kova{\v{c}}i{\v{c}}]{pervs2010histograms}
Janez Per{\v{s}}, Vildana Suli{\'c}, Matej Kristan, Matej Per{\v{s}}e, Klemen Polanec, and Stanislav Kova{\v{c}}i{\v{c}}.
\newblock Histograms of optical flow for efficient representation of body motion.
\newblock \emph{Pattern Recognition Letters}, 31\penalty0 (11):\penalty0 1369--1376, 2010.

\bibitem[Poppe(2010)]{poppe2010survey}
Ronald Poppe.
\newblock A survey on vision-based human action recognition.
\newblock \emph{Image and vision computing}, 28\penalty0 (6):\penalty0 976--990, 2010.

\bibitem[Radford et~al.(2021)Radford, Kim, Hallacy, Ramesh, Goh, Agarwal, Sastry, Askell, Mishkin, Clark, et~al.]{radford2021learning}
Alec Radford, Jong~Wook Kim, Chris Hallacy, Aditya Ramesh, Gabriel Goh, Sandhini Agarwal, Girish Sastry, Amanda Askell, Pamela Mishkin, Jack Clark, et~al.
\newblock Learning transferable visual models from natural language supervision.
\newblock In \emph{International conference on machine learning}, pages 8748--8763. PmLR, 2021.

\bibitem[Rambhatla et~al.(2023)Rambhatla, Misra, Chellappa, and Shrivastava]{Rambhatla2023MOSTMO}
Sai~Saketh Rambhatla, Ishan Misra, Rama Chellappa, and Abhinav Shrivastava.
\newblock Most: Multiple object localization with self-supervised transformers for object discovery.
\newblock \emph{2023 IEEE/CVF International Conference on Computer Vision (ICCV)}, pages 15777--15788, 2023.

\bibitem[Shao et~al.(2020)Shao, Zhao, Dai, and Lin]{finegym}
Dian Shao, Yue Zhao, Bo Dai, and Dahua Lin.
\newblock Finegym: A hierarchical video dataset for fine-grained action understanding.
\newblock In \emph{IEEE Conference on Computer Vision and Pattern Recognition (CVPR)}, 2020.

\bibitem[Sim\'eoni et~al.(2021)Sim\'eoni, Puy, Vo, Roburin, Gidaris, Bursuc, P\'erez, Marlet, and Ponce]{LOST}
Oriane Sim\'eoni, Gilles Puy, Huy~V. Vo, Simon Roburin, Spyros Gidaris, Andrei Bursuc, Patrick P\'erez, Renaud Marlet, and Jean Ponce.
\newblock Localizing objects with self-supervised transformers and no labels.
\newblock 2021.

\bibitem[Simonyan and Zisserman(2014)]{simonyan2014two}
Karen Simonyan and Andrew Zisserman.
\newblock Two-stream convolutional networks for action recognition in videos.
\newblock \emph{Advances in neural information processing systems}, 27, 2014.

\bibitem[Soomro et~al.(2012)Soomro, Zamir, and Shah]{ucf}
Khurram Soomro, Amir Zamir, and Mubarak Shah.
\newblock Ucf101: A dataset of 101 human actions classes from videos in the wild.
\newblock \emph{ArXiv}, abs/1212.0402, 2012.

\bibitem[Tang et~al.(2024)Tang, B{\'e}jar, and Vidal]{tang2024semantic}
Yutao Tang, Benjam{\'\i}n B{\'e}jar, and Ren{\'e} Vidal.
\newblock Semantic-aware video representation for few-shot action recognition.
\newblock In \emph{Proceedings of the IEEE/CVF Winter Conference on Applications of Computer Vision}, pages 6458--6468, 2024.

\bibitem[Thatipelli et~al.(2022)Thatipelli, Narayan, Khan, Anwer, Khan, and Ghanem]{strm}
Anirudh Thatipelli, Sanath Narayan, Salman Khan, Rao~Muhammad Anwer, Fahad~Shahbaz Khan, and Bernard Ghanem.
\newblock Spatio-temporal relation modeling for few-shot action recognition.
\newblock In \emph{Proceedings of the IEEE/CVF Conference on Computer Vision and Pattern Recognition}, pages 19958--19967, 2022.

\bibitem[Tran et~al.(2015)Tran, Bourdev, Fergus, Torresani, and Paluri]{tran2015learning}
Du Tran, Lubomir Bourdev, Rob Fergus, Lorenzo Torresani, and Manohar Paluri.
\newblock Learning spatiotemporal features with 3d convolutional networks.
\newblock In \emph{Proceedings of the IEEE international conference on computer vision}, pages 4489--4497, 2015.

\bibitem[Tumanyan et~al.(2024)Tumanyan, Singer, Bagon, and Dekel]{dino_tracker}
Narek Tumanyan, Assaf Singer, Shai Bagon, and Tali Dekel.
\newblock Dino-tracker: Taming dino for self-supervised point tracking in a single video, 2024.

\bibitem[Wang and Schmid(2013)]{wang2013action}
Heng Wang and Cordelia Schmid.
\newblock Action recognition with improved trajectories.
\newblock In \emph{Proceedings of the IEEE international conference on computer vision}, pages 3551--3558, 2013.

\bibitem[Wang et~al.(2013)Wang, Kl{\"a}ser, Schmid, and Liu]{wang2013dense}
Heng Wang, Alexander Kl{\"a}ser, Cordelia Schmid, and Cheng-Lin Liu.
\newblock Dense trajectories and motion boundary descriptors for action recognition.
\newblock \emph{International journal of computer vision}, 103:\penalty0 60--79, 2013.

\bibitem[Wang et~al.(2020)Wang, Tran, Torresani, and Feiszli]{wang2020video}
Heng Wang, Du Tran, Lorenzo Torresani, and Matt Feiszli.
\newblock Video modeling with correlation networks.
\newblock In \emph{Proceedings of the IEEE/CVF Conference on Computer Vision and Pattern Recognition}, pages 352--361, 2020.

\bibitem[Wang et~al.(2025)Wang, Chen, Karaev, Vedaldi, Rupprecht, and Novotny]{vggt}
Jianyuan Wang, Minghao Chen, Nikita Karaev, Andrea Vedaldi, Christian Rupprecht, and David Novotny.
\newblock Vggt: Visual geometry grounded transformer.
\newblock In \emph{Proceedings of the Computer Vision and Pattern Recognition Conference}, pages 5294--5306, 2025.

\bibitem[Wang et~al.(2015)Wang, Qiao, and Tang]{wang2015action}
Limin Wang, Yu Qiao, and Xiaoou Tang.
\newblock Action recognition with trajectory-pooled deep-convolutional descriptors.
\newblock In \emph{Proceedings of the IEEE conference on computer vision and pattern recognition}, pages 4305--4314, 2015.

\bibitem[Wang et~al.(2018)Wang, Li, Li, and Van~Gool]{wang2018appearance}
Limin Wang, Wei Li, Wen Li, and Luc Van~Gool.
\newblock Appearance-and-relation networks for video classification.
\newblock In \emph{Proceedings of the IEEE conference on computer vision and pattern recognition}, pages 1430--1439, 2018.

\bibitem[Wang et~al.(2016)Wang, Li, Hou, and Li]{wang2016action}
Pichao Wang, Zhaoyang Li, Yonghong Hou, and Wanqing Li.
\newblock Action recognition based on joint trajectory maps using convolutional neural networks.
\newblock In \emph{Proceedings of the 24th ACM international conference on Multimedia}, pages 102--106, 2016.

\bibitem[Wang et~al.(2023{\natexlab{a}})Wang, Chang, Cai, Li, Hariharan, Holynski, and Snavely]{wang2023tracking}
Qianqian Wang, Yen-Yu Chang, Ruojin Cai, Zhengqi Li, Bharath Hariharan, Aleksander Holynski, and Noah Snavely.
\newblock Tracking everything everywhere all at once.
\newblock \emph{arXiv preprint arXiv:2306.05422}, 2023{\natexlab{a}}.

\bibitem[Wang et~al.(2021{\natexlab{a}})Wang, Qing, Huang, Feng, Zhang, Jiang, Tang, Gao, and Sang]{wang2021proposal}
Xiang Wang, Zhiwu Qing, Ziyuan Huang, Yutong Feng, Shiwei Zhang, Jianwen Jiang, Mingqian Tang, Changxin Gao, and Nong Sang.
\newblock Proposal relation network for temporal action detection.
\newblock \emph{arXiv preprint arXiv:2106.11812}, 2021{\natexlab{a}}.

\bibitem[Wang et~al.(2021{\natexlab{b}})Wang, Ye, Qi, Zhao, Wang, Shan, and Wang]{wang2021semantic}
Xiao Wang, Weirong Ye, Zhongang Qi, Xun Zhao, Guangge Wang, Ying Shan, and Hanzi Wang.
\newblock Semantic-guided relation propagation network for few-shot action recognition.
\newblock In \emph{Proceedings of the 29th ACM International Conference on Multimedia}, pages 816--825, 2021{\natexlab{b}}.

\bibitem[Wang et~al.(2021{\natexlab{c}})Wang, Zhang, Qing, Shao, Gao, and Sang]{wang2021self}
Xiang Wang, Shiwei Zhang, Zhiwu Qing, Yuanjie Shao, Changxin Gao, and Nong Sang.
\newblock Self-supervised learning for semi-supervised temporal action proposal.
\newblock In \emph{Proceedings of the IEEE/CVF Conference on Computer Vision and Pattern Recognition}, pages 1905--1914, 2021{\natexlab{c}}.

\bibitem[Wang et~al.(2022)Wang, Zhang, Qing, Tang, Zuo, Gao, Jin, and Sang]{hyrsm}
Xiang Wang, Shiwei Zhang, Zhiwu Qing, Mingqian Tang, Zhengrong Zuo, Changxin Gao, Rong Jin, and Nong Sang.
\newblock Hybrid relation guided set matching for few-shot action recognition.
\newblock In \emph{Proceedings of the IEEE/CVF Conference on Computer Vision and Pattern Recognition}, pages 19948--19957, 2022.

\bibitem[Wang et~al.(2023{\natexlab{b}})Wang, Girdhar, Yu, and Misra]{Wang2023CutAL}
Xudong Wang, Rohit Girdhar, Stella~X. Yu, and Ishan Misra.
\newblock Cut and learn for unsupervised object detection and instance segmentation.
\newblock \emph{2023 IEEE/CVF Conference on Computer Vision and Pattern Recognition (CVPR)}, pages 3124--3134, 2023{\natexlab{b}}.

\bibitem[Wang et~al.(2023{\natexlab{c}})Wang, Misra, Zeng, Girdhar, and Darrell]{Wang2023VideoCutLERSS}
Xudong Wang, Ishan Misra, Ziyun Zeng, Rohit Girdhar, and Trevor Darrell.
\newblock Videocutler: Surprisingly simple unsupervised video instance segmentation.
\newblock \emph{2024 IEEE/CVF Conference on Computer Vision and Pattern Recognition (CVPR)}, pages 22755--22764, 2023{\natexlab{c}}.

\bibitem[Wang et~al.(2023{\natexlab{d}})Wang, Zhang, Qing, Gao, Zhang, Zhao, and Sang]{molo}
Xiang Wang, Shiwei Zhang, Zhiwu Qing, Changxin Gao, Yingya Zhang, Deli Zhao, and Nong Sang.
\newblock Molo: Motion-augmented long-short contrastive learning for few-shot action recognition.
\newblock In \emph{Proceedings of the IEEE/CVF Conference on Computer Vision and Pattern Recognition}, pages 18011--18021, 2023{\natexlab{d}}.

\bibitem[Wang et~al.(2024{\natexlab{a}})Wang, Yan, Hu, Li, and Wang]{ccln}
Xiao Wang, Yan Yan, Hai-Miao Hu, Bo Li, and Hanzi Wang.
\newblock Cross-modal contrastive learning network for few-shot action recognition.
\newblock \emph{IEEE Transactions on Image Processing}, 2024{\natexlab{a}}.

\bibitem[Wang et~al.(2024{\natexlab{b}})Wang, Zhang, Cen, Gao, Zhang, Zhao, and Sang]{wang2024clip}
Xiang Wang, Shiwei Zhang, Jun Cen, Changxin Gao, Yingya Zhang, Deli Zhao, and Nong Sang.
\newblock Clip-guided prototype modulating for few-shot action recognition.
\newblock \emph{International Journal of Computer Vision}, 132\penalty0 (6):\penalty0 1899--1912, 2024{\natexlab{b}}.

\bibitem[Wang et~al.(2024{\natexlab{c}})Wang, Zhang, Qing, Zuo, Gao, Jin, and Sang]{hyrsm++}
Xiang Wang, Shiwei Zhang, Zhiwu Qing, Zhengrong Zuo, Changxin Gao, Rong Jin, and Nong Sang.
\newblock Hyrsm++: Hybrid relation guided temporal set matching for few-shot action recognition.
\newblock \emph{Pattern Recognition}, 147:\penalty0 110110, 2024{\natexlab{c}}.

\bibitem[Wang et~al.(2024{\natexlab{d}})Wang, Guo, Zhu, and Guo]{wang2024sfmm}
Yikang Wang, Fei Guo, Li Zhu, and Yuan Guo.
\newblock Sfmm: Semantic-to-frame matching with multi-classifier for few-shot action recognition.
\newblock In \emph{2024 International Joint Conference on Neural Networks (IJCNN)}, pages 1--8. IEEE, 2024{\natexlab{d}}.

\bibitem[Wu et~al.(2024{\natexlab{a}})Wu, Wu, Li, Xu, Feng, and Kittler]{wu2024efficient}
Cong Wu, Xiao-Jun Wu, Linze Li, Tianyang Xu, Zhenhua Feng, and Josef Kittler.
\newblock Efficient few-shot action recognition via multi-level post-reasoning.
\newblock In \emph{European Conference on Computer Vision}, pages 38--56. Springer, 2024{\natexlab{a}}.

\bibitem[Wu et~al.(2022)Wu, Zhang, Zhang, Wu, and Zhang]{mtfan}
Jiamin Wu, Tianzhu Zhang, Zhe Zhang, Feng Wu, and Yongdong Zhang.
\newblock Motion-modulated temporal fragment alignment network for few-shot action recognition.
\newblock In \emph{Proceedings of the IEEE/CVF Conference on Computer Vision and Pattern Recognition}, pages 9151--9160, 2022.

\bibitem[Wu et~al.(2024{\natexlab{b}})Wu, Guan, Li, Huang, Liu, Wang, Xian, Shrivastava, Huang, Boyd-Graber, et~al.]{wu2024autohallusion}
Xiyang Wu, Tianrui Guan, Dianqi Li, Shuaiyi Huang, Xiaoyu Liu, Xijun Wang, Ruiqi Xian, Abhinav Shrivastava, Furong Huang, Jordan~Lee Boyd-Graber, et~al.
\newblock Autohallusion: Automatic generation of hallucination benchmarks for vision-language models.
\newblock \emph{arXiv preprint arXiv:2406.10900}, 2024{\natexlab{b}}.

\bibitem[Xia et~al.(2023)Xia, Li, Min, and Ding]{rfpl}
Haifeng Xia, Kai Li, Martin~Renqiang Min, and Zhengming Ding.
\newblock Few-shot video classification via representation fusion and promotion learning.
\newblock In \emph{Proceedings of the IEEE/CVF international conference on computer vision}, pages 19311--19320, 2023.

\bibitem[Xiao et~al.(2024)Xiao, Wang, Zhang, Xue, Peng, Shen, and Zhou]{SpatialTracker}
Yuxi Xiao, Qianqian Wang, Shangzhan Zhang, Nan Xue, Sida Peng, Yujun Shen, and Xiaowei Zhou.
\newblock Spatialtracker: Tracking any 2d pixels in 3d space.
\newblock In \emph{Proceedings of the IEEE/CVF Conference on Computer Vision and Pattern Recognition (CVPR)}, 2024.

\bibitem[Xiao et~al.(2025)Xiao, Wang, Xue, Karaev, Makarov, Kang, Zhu, Bao, Shen, and Zhou]{xiao2025spatialtrackerv23dpointtracking}
Yuxi Xiao, Jianyuan Wang, Nan Xue, Nikita Karaev, Yuri Makarov, Bingyi Kang, Xing Zhu, Hujun Bao, Yujun Shen, and Xiaowei Zhou.
\newblock Spatialtrackerv2: 3d point tracking made easy, 2025.

\bibitem[Xing et~al.(2023{\natexlab{a}})Xing, Wang, Mu, and Liu]{sloshnet}
Jiazheng Xing, Mengmeng Wang, Boyu Mu, and Yong Liu.
\newblock Revisiting the spatial and temporal modeling for few-shot action recognition.
\newblock In \emph{AAAI Conference on Artificial Intelligence}, 2023{\natexlab{a}}.

\bibitem[Xing et~al.(2023{\natexlab{b}})Xing, Wang, Ruan, Chen, Guo, Mu, Dai, Wang, and Liu]{gghm}
Jiazheng Xing, Mengmeng Wang, Yudi Ruan, Bofan Chen, Yaowei Guo, Boyu Mu, Guang Dai, Jingdong Wang, and Yong Liu.
\newblock Boosting few-shot action recognition with graph-guided hybrid matching.
\newblock In \emph{Proceedings of the IEEE/CVF International Conference on Computer Vision}, pages 1740--1750, 2023{\natexlab{b}}.

\bibitem[Zhang et~al.(2025)Zhang, Ke, Harley, and Fragkiadaki]{zhang2025tapip3d}
Bowei Zhang, Lei Ke, Adam~W Harley, and Katerina Fragkiadaki.
\newblock Tapip3d: Tracking any point in persistent 3d geometry.
\newblock \emph{arXiv preprint arXiv:2504.14717}, 2025.

\bibitem[Zhang et~al.(2023)Zhang, Zhu, Wang, Chen, Wu, and Wang]{zhang2023extracting}
Guozhen Zhang, Yuhan Zhu, Haonan Wang, Youxin Chen, Gangshan Wu, and Limin Wang.
\newblock Extracting motion and appearance via inter-frame attention for efficient video frame interpolation.
\newblock In \emph{Proceedings of the IEEE/CVF Conference on Computer Vision and Pattern Recognition}, pages 5682--5692, 2023.

\bibitem[Zhang et~al.(2020)Zhang, Zhang, Qi, Li, Torr, and Koniusz]{zhang2020few}
Hongguang Zhang, Li Zhang, Xiaojuan Qi, Hongdong Li, Philip~HS Torr, and Piotr Koniusz.
\newblock Few-shot action recognition with permutation-invariant attention.
\newblock In \emph{Computer Vision--ECCV 2020: 16th European Conference, Glasgow, UK, August 23--28, 2020, Proceedings, Part V 16}, pages 525--542. Springer, 2020.

\bibitem[Zhang et~al.(2021)Zhang, Zhou, and He]{zhang2021learning}
Songyang Zhang, Jiale Zhou, and Xuming He.
\newblock Learning implicit temporal alignment for few-shot video classification.
\newblock \emph{arXiv preprint arXiv:2105.04823}, 2021.

\bibitem[Zhao et~al.(2018)Zhao, Xiong, and Lin]{zhao2018recognize}
Yue Zhao, Yuanjun Xiong, and Dahua Lin.
\newblock Recognize actions by disentangling components of dynamics.
\newblock In \emph{Proceedings of the IEEE Conference on Computer Vision and Pattern Recognition}, pages 6566--6575, 2018.

\bibitem[Zheng et~al.(2024)Zheng, Liang, Huang, Gao, Daum{\'e}~III, Kolobov, Huang, and Yang]{zheng2024tracevla}
Ruijie Zheng, Yongyuan Liang, Shuaiyi Huang, Jianfeng Gao, Hal Daum{\'e}~III, Andrey Kolobov, Furong Huang, and Jianwei Yang.
\newblock Tracevla: Visual trace prompting enhances spatial-temporal awareness for generalist robotic policies.
\newblock \emph{arXiv preprint arXiv:2412.10345}, 2024.

\bibitem[Zheng et~al.(2022)Zheng, Chen, and Jin]{hcl}
Sipeng Zheng, Shizhe Chen, and Qin Jin.
\newblock Few-shot action recognition with hierarchical matching and contrastive learning.
\newblock In \emph{European Conference on Computer Vision}, pages 297--313. Springer, 2022.

\bibitem[Zheng et~al.(2023)Zheng, Harley, Shen, Wetzstein, and Guibas]{zheng2023pointodyssey}
Yang Zheng, Adam~W Harley, Bokui Shen, Gordon Wetzstein, and Leonidas~J Guibas.
\newblock Pointodyssey: A large-scale synthetic dataset for long-term point tracking.
\newblock In \emph{Proceedings of the IEEE/CVF International Conference on Computer Vision}, pages 19855--19865, 2023.

\bibitem[Zholus et~al.(2025)Zholus, Doersch, Yang, Koppula, Patraucean, He, Rocco, Sajjadi, Chandar, and Goroshin]{tapnext}
Artem Zholus, Carl Doersch, Yi Yang, Skanda Koppula, Viorica Patraucean, Xu~Owen He, Ignacio Rocco, Mehdi~SM Sajjadi, Sarath Chandar, and Ross Goroshin.
\newblock Tapnext: Tracking any point (tap) as next token prediction.
\newblock \emph{arXiv preprint arXiv:2504.05579}, 2025.

\bibitem[Zhu et~al.(2017)Zhu, Zhao, Huang, Tu, and Ma]{zhu2017structured}
Chen Zhu, Yanpeng Zhao, Shuaiyi Huang, Kewei Tu, and Yi Ma.
\newblock Structured attentions for visual question answering.
\newblock In \emph{Proceedings of the IEEE International Conference on Computer Vision}, pages 1291--1300, 2017.

\bibitem[Zhu and Yang(2018)]{zhu2018compound}
Linchao Zhu and Yi Yang.
\newblock Compound memory networks for few-shot video classification.
\newblock In \emph{Proceedings of the European Conference on Computer Vision (ECCV)}, pages 751--766, 2018.

\bibitem[Zhu and Yang(2020)]{zhu2020label}
Linchao Zhu and Yi Yang.
\newblock Label independent memory for semi-supervised few-shot video classification.
\newblock \emph{IEEE Transactions on Pattern Analysis and Machine Intelligence}, 44\penalty0 (1):\penalty0 273--285, 2020.

\end{thebibliography}
}

\end{document}